\documentclass[a4paper,fleqn]{cas-dc}

\usepackage{graphicx}
\usepackage{amsmath}
\usepackage{caption}
\usepackage{multirow}
\usepackage{amssymb}
\usepackage[caption=false]{subfig}

\usepackage[numbers]{natbib}
\def\tsc#1{\csdef{#1}{\textsc{\lowercase{#1}}\xspace}}
\tsc{WGM}
\tsc{QE}
\begin{document}
\let\WriteBookmarks\relax
\def\floatpagepagefraction{1}
\def\textpagefraction{.001}

\shorttitle{Ins-ATP: Deep Estimation of ATP for Organoid}    

\shortauthors{Xs. Bian et~al.}  

\title [mode = title]{Ins-ATP: Deep Estimation of ATP for Organoid Based on High Throughput Microscopic Images}  

%\tnotemark[<tnote number>] 

%\tnotetext[<tnote number>]{<tnote text>} 


\author[1,2]{Xuesheng Bian}[
		orcid=0000-0002-7690-4386,
]

%\fnmark[]

\ead{xsbian@ycit.edu.cn}
	
%\ead[url]{<URL>}

\credit{Conceptualization, Methodology, Writing - Original Draft}

\author[2]{Cheng Wang}[]
\ead{cwang@xmu.edu.cn}
\cormark[1]
\credit{Conceptualization, Methodology, Supervision, Resources}

\author[5]{Shuting Chen}[]
\ead{chenst2016@jmu.edu.cn}
\credit{Resources, Funding acquisition}

\author[2]{Weiquan Liu}[]
\ead{wqliu@xmu.edu.cn}
\credit{Methodology, Writing - Review \& Editing, Funding acquisition}

\author[1]{Sen Xu}[]
\ead{xusen@ycit.cn}
\credit{Funding acquisition}

\author[1]{Jinxin Zhu}[]
\ead{zjx@ycit.edu.cn}
\credit{Resources}

\author[1]{Rugang Wang}[]
\ead{wrg3506@ycit.edu.cn}
\credit{Resources}

\author[4]{Zexin Chen}[]
\ead{czxchase@126.com}
\credit{Specimen culture, Resources}

\author[4]{Min Huang}[]
\ead{huangmin@accibio.com}
\credit{Investigation, Resources}

\author[3]{Gang Li}[]
\ead{lg@smu.edu.cn}
\cormark[1]
\credit{Conceptualization, Methodology, Resources}

\affiliation[1]{organization={Yancheng Institute of Technology},
	addressline={Jiangsu province Yancheng City Hope Avenue Road 1}, 
	city={Yancheng},
	citysep={}, % Uncomment if no comma needed between city and postcode
	postcode={224051}, 
	state={Jiangsu},
	country={China}}

\affiliation[2]{organization={Fujian Key Laboratory of Sensing and Computing for Smart Cities, School of Informatics, Xiamen University},
	addressline={Fujian province Xiamen City Siming South Road No. 422}, 
	city={Xiamen},
	citysep={}, % Uncomment if no comma needed between city and postcode
	postcode={361005}, 
	state={Fujian},
	country={China}}

\affiliation[3]{organization={Department of Otolaryngology-Head and Neck Surgery, Precision Medical Center, Nanfang Hospital},
	addressline={Guangdong province Guangzhou City Baiyun District North Guangzhou Avenue No. 1838}, 
	city={Guangzhou},
	citysep={}, % Uncomment if no comma needed between city and postcode
	postcode={510000}, 
	state={Guangdong},
	country={China}}

\affiliation[4]{organization={Guangdong Research Center of Organoid Engineering and Technology},
	addressline={Guangdong province Guangzhou City Huangpu District Kaiyuan Avenue No. 11}, 
	city={Guangzhou},
	citysep={}, % Uncomment if no comma needed between city and postcode
	postcode={510000}, 
	state={Guangdong},
	country={China}}

\affiliation[5]{organization={Chengyi University College, Jimei University},
	addressline={Fujian province Xiamen City Jimei District Jimei Boulevard 199}, 
	city={Xiamen},
	citysep={}, % Uncomment if no comma needed between city and postcode
	postcode={361021}, 
	state={Fujian},
	country={China}}

% \author[<aff no>]{<author name>}[<options>]

% Footnote of the second author

% Email id of the second author
% \ead{}



%\affiliation[<aff no>]{organization={},
%          citysep={}, % Uncomment if no comma needed between city and postcode

%\cortext[1]{Corresponding author}



\begin{abstract}
Adenosine triphosphate (ATP) is a high-energy phosphate compound and the most direct energy source in organisms. ATP is an essential biomarker for evaluating cell viability in biology. Researchers often use ATP bioluminescence to measure the ATP of organoid after drug to evaluate the drug efficacy. However, ATP bioluminescence has some limitations, leading to unreliable drug screening results. Performing ATP bioluminescence causes cell lysis of organoids, so it is impossible to observe organoids' long-term viability changes after medication continually. To overcome the disadvantages of ATP bioluminescence, we propose Ins-ATP, a non-invasive strategy, the first organoid ATP estimation model based on the high-throughput microscopic image. Ins-ATP directly estimates the ATP of organoids from high-throughput microscopic images, so that it does not influence the drug reactions of organoids. Therefore, the ATP change of organoids can be observed for a long time to obtain more stable results. Experimental results show that the ATP estimation by Ins-ATP is in good agreement with those determined by ATP bioluminescence. Specifically, the predictions of Ins-ATP are consistent with the results measured by ATP bioluminescence in the efficacy evaluation experiments of different drugs.
\end{abstract}


%\item We construct the first organoid ATP estimation dataset, which is vital for organoid biomarker estimation.
%\item To our knowledge, this method is the first exploration of applying deep learning to ATP estimation for organoids.
%\item Experiments show that the proposed Ins-ATP achieves accurate ATP estimation of organoids, and its predictions are in high agreement with ATP bioluminescence. In addition, Ins-ATP's evaluation of the efficacy differences of different drugs has a high correlation with the results of ATP bioluminescence.

% Keywords

\maketitle

% Main text
\section{Introduction}\label{}
In biopharmaceuticals, cell models and animal models are the mainstream models for drug screening. The culture of the cell model is easy, low cost, and the culture process can be precisely controlled, which enables large-scale drug screening \cite{sterneckert2014investigating}. However, the structure of the cell model is simple, making it unable to simulate the accurate response of drugs in a complex environment in vivo \cite{munoz2020animal}. Compared with cell models, animal models can better simulate the complex environment of drugs in vivo, and the results of drug experiments are more convincing. However, animal models are high-cost and difficult to conduct large-scale experiments. 

Organoids have been a research hotspot in recent years. It is an ideal drug test model, taking advantage of both cell models and animal models \cite{park2019organoids,tuveson2019cancer}. Organoids are 3D cell complexes cultured in vitro with highly similar structures and functions to the source organ or tissue. Organoids can be long-term cultured in vitro and have stable phenotypic and genetic characteristics. Compared with traditional 2D cell culture methods, organoids contain complex cell types and more complex biological communication. Therefore, organoids can better simulate the growth and development process and physiological and pathological states of organs or tissues, playing a significant role in drug screening and personalized medicine.

At present, researchers worldwide have made many efforts in organoid culture. They have successfully cultured various organoids, including normal organs such as the liver, kidney, stomach, heart, and various tumour organoids, e.g. breast cancer, lung cancer, and pancreatic cancer \cite{park2019organoids}. Organoid culture technology is developing rapidly, and various types of organoids are gradually successfully cultured. However, organoid-related analysis and processing technologies still rely on traditional methods designed for cell models, and there are few significant breakthroughs.

Microscopic imaging is essential for organoid analysis. Researchers use optical microscopes to observe the morphology and growth trends of organoids. However, organoid biomarkers, critical in organoid research, are difficult to estimate using microscopic images. Researchers have to use various reagents or professional instruments to measure the biomarkers of organoids. This process changes the original growth environment of the organoids and causes irreversible effects on the organoids. 

Organoid microscopic images contain quite a wealth of information. After medication, drugs with superior efficacy have a massive impact on the development of organoids. We observed that the organoids show different appearances after medication, such as decomposition, contraction, and ablation. Specifically, obtaining microscopic images of organoids does not affect the growth of the organoids and interfere with the effect of drugs on the organoids. Therefore, if the discriminative features related to organoid biomarkers can be extracted from organoid microscope images, the estimation from microscope images to organoid biomarkers can be achieved.

Numerous studies show that deep learning technology represented by deep neural networks (DNNs) has great potential in computer vision \cite{litjens2017survey,pouyanfar2018survey}. After years of development, deep learning technology has penetrated various computer vision tasks, such as object recognition \cite{resnet,vgg,inception,googlenet}, object detection \cite{carion2020end,redmon2016you,ren2015faster}, segmentation \cite{chen2017deeplab,ronneberger2015u,wang2021max}, super-resolution \cite{dong2015image,dong2016accelerating,lu2022transformer}, registration \cite{,chen2022transmorph,haskins2020deep,liu2018h}, etc. Deep neural networks (DNNs) effectively learn important features in images and improve the representation ability of image features through end-to-end learning. Therefore, we hope to estimate organoid biomarkers from microscopic images of organoids through deep learning techniques. 

In organoid drug screening experiments, ATP bioluminescence is an important means to assess the viability of organoids \cite{sevin1988application}. However, ATP bioluminescence causes organoid lysis, making it hard to obtain the changing trend of organoid viability after performing ATP bioluminescence. Therefore, performing ATP bioluminescence at different times may result in different ATP measurement results, making organoid drug screening results unreliable. There are few studies on non-invasive biomarkers estimation for organoids through investigation. In this paper, we take the ATP of organoids as an example to explore the application potential of deep learning techniques in organoid drug screening tasks. 

Organoid ATP estimation using deep learning is a challenging task. First, deep learning techniques rely on a large number of labelled samples, but there are currently no open-source datasets for this task. Second, only living cells contain ATP, and organoid images are filled with many ATP-independent targets, such as vacuoles, impurities, background, etc. These objectives easily cause instability of the model optimization during the training process, resulting in disturbance of the estimation results. Finally, the ATP range of organoids is large, and the samples distribute unevenly, which causes the model prediction results to be inaccurate for extreme samples.

In this study, we propose a multi-instance learning based model, Ins-ATP, to address the above challenges. Ins-ATP accurately estimates the ATP of organoids from microscopic images. To our knowledge, this is the first attempt that introduces deep learning to solve ATP estimation for organoids. The contribution of the method proposed in this paper are as follows:

\begin{itemize}
	\item We construct the first organoid ATP estimation dataset, which is vital for organoid biomarker estimation.
	\item We design the first organoid ATP estimation model, Ins-ATP. We introduce a multi-instance learning strategy, which effectively overcomes the disturbance of objects unrelated to ATP estimation in the image and improves the prediction accuracy and robustness.
	\item The proposed Ins-ATP achieves accurate ATP estimation of organoids, and its predictions are in high agreement with ATP bioluminescence. In addition, Ins-ATP's evaluation of the efficacy differences of different drugs has a high correlation with the results of ATP bioluminescence.
\end{itemize}

% Use the style of numbering in square brackets.
% If nothing is used, default style will be taken.
%\begin{enumerate}[a)]
%\item 
%\item 
%\item 
%\end{enumerate}  

%\begin{itemize}
%\item 
%\item 
%\item 
%\end{itemize}  

%\begin{description}
%\item[]
%\item[] 
%\item[] 
%\end{description}  

% Figure
%\begin{figure}[<options>]
%	\centering
%		\includegraphics[<options>]{}
%	  \caption{}\label{fig1}
%\end{figure}

%\begin{table}[<options>]
%\caption{}\label{tbl1}
%\begin{tabular*}{\tblwidth}{@{}LL@{}}
%\toprule
%  &  \\ % Table header row
%\midrule
% & \\
% & \\
% & \\
% & \\
%\bottomrule
%\end{tabular*}
%\end{table}

% Uncomment and use as the case may be
%\begin{theorem} 
%\end{theorem}

% Uncomment and use as the case may be
%\begin{lemma} 
%\end{lemma}


\section{Related works}\label{}

Organoid viability evaluation is a critical step in organoid drug screening. Conventional ATP bioluminescence interferes with the normal development of the organoid. Therefore, researchers have been exploring easy-to-use and non-invasive methods for organoid viability evaluation. Some existing works have initially explored organoid viability evaluation using organoid imaging. The challenge in achieving organoid viability evaluation by organoid imaging is the high amount of redundant information in organoid images. 

Thompson et al. \cite{thompson2022evaluation} proposed to evaluate organoid viability by counting the change in the number of organoids after thawing and the percentage of live cells after fluorescent staining. This method considered microscopic observation to evaluate organoid viability and reduce the impact on the sample. However, this is only a qualitative assessment method, and its accuracy cannot be quantified. Powell et al. \cite{powell2022deeporganoid} proposed an AutoML-based deep learning model to regress the terminal viability readout (CellTiter-Glo) in high-throughput microscopic images and achieved high performance. However, the method did not address the problem of highly redundant information in high-throughput microscopic images in a targeted manner. Thus the correlation between the predicted and true values on different sample groups is unstable. 

In the previous works, we proposed an organoid viability evaluation method based on the expert weak label before, which combines organoid detection and organoid viability scoring to achieve the evaluation of whole-well sample viability \cite{bian2021deep,bian2021organet}. However, this method performed detection in advance to reduce the interference of redundant information in images but focused mainly on the larger-sized organoids in 96-well plates and was not compatible to samples in 384-well plates.

Since ATP of the organoid is only strongly correlated with the organoid, the target changes in the microscopic images that are unrelated to ATP should not impact the ATP estimation results. However, deep learning methods are more sensitive to changes in the images, and slight changes in the images may cause severe perturbations in the prediction results. Therefore, the key to improving organoid ATP estimation performance lies in overcoming the discrepancy of prediction results caused by the perturbation of targets other than organoids in the images. We find that the characteristics of the organoid ATP estimation task are highly compatible with the multi-instance learning paradigm\cite{zhou2004multi}. 

In the multi-instance learning approach, the basic unit of training samples is the bag. Each bag contains multiple samples, and the samples in the bag are called instances, and each instance has a different impact on the bag according to its importance. Multi-instance learning has many applications in pathological image analysis tasks that similar to high-throughput organoid microscopic image analysis. Campanell et al. \cite{campanella2019clinical} proposed a multi-instance learning based approach that uses diagnostic reports as training labels to avoid expensive and time-consuming manual annotation of pixels. Naik et al. \cite{naik2020deep} proposed a multi-instance based approach to directly assess Estrogen Receptor Status (ERS) from pathology images. Lerousseau et al. \cite{lerousseau2020weakly} proposed a multi-scale domain adversarial multi-instance convolutional network for cancer subtype classification. All of the above methods used multi-instance learning to overcome the problem of too much redundant information in pathology images.

\section{Material and methods}\label{}
\subsection{Data}
\begin{figure*}[t]
	\centering
	\subfloat[Historgam for ATP of tumor organoid]{\includegraphics[width=0.48\textwidth]{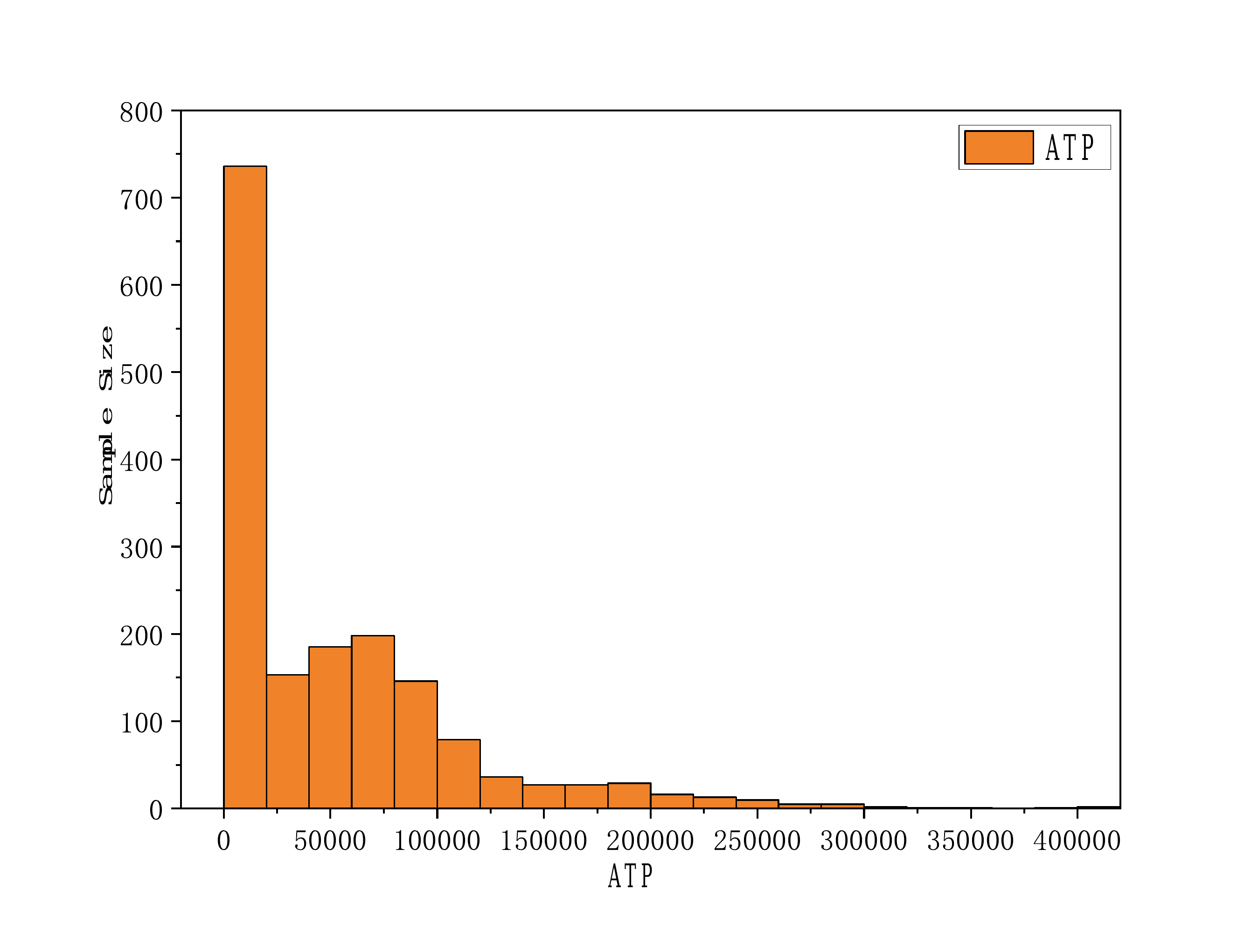}
		\label{fig:samples:a}}
	\subfloat[ATP statistics for tumor organoid]{\includegraphics[width=0.48\textwidth]{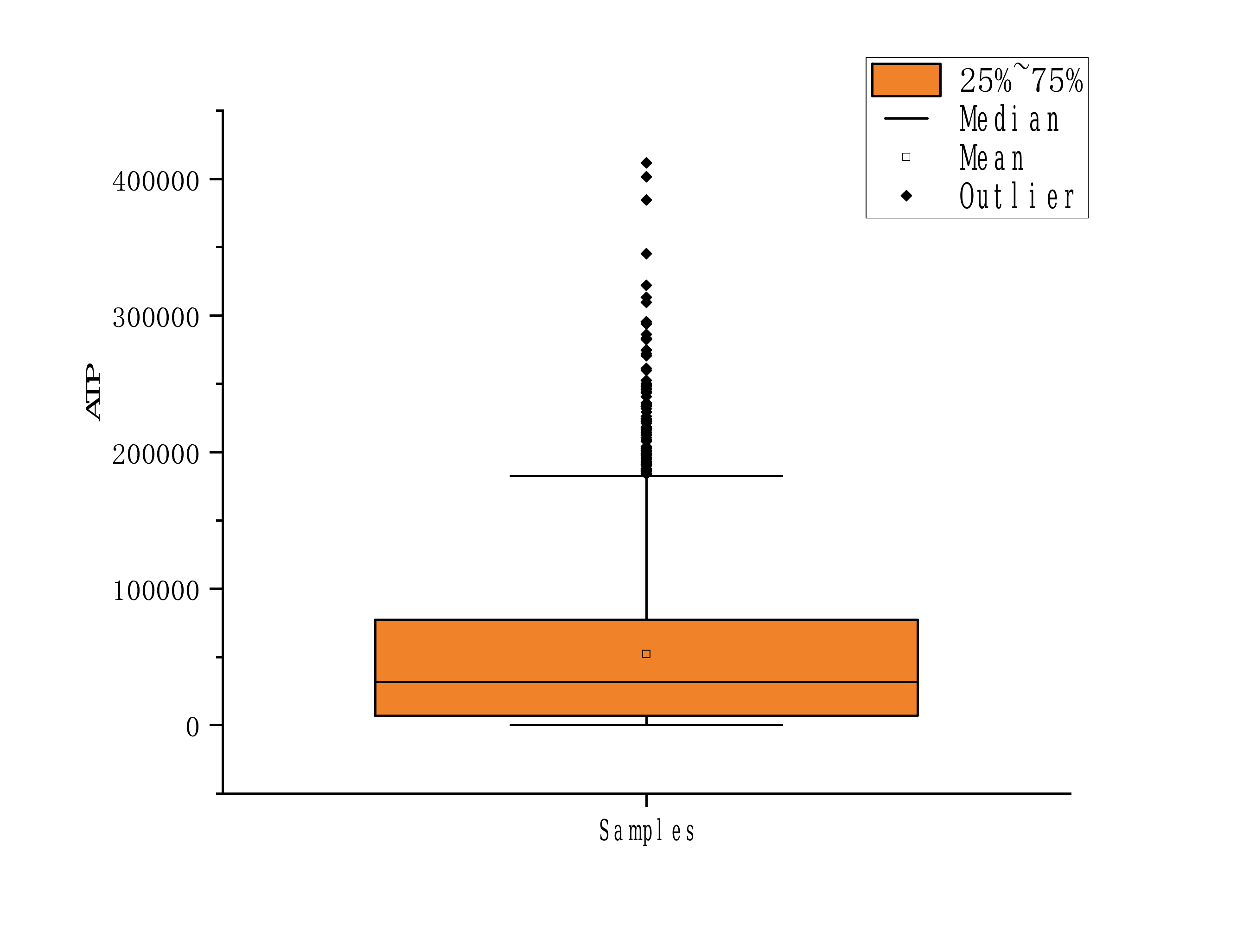}
		\label{fig:samples:b}}
	\caption{The samples distributed unevenly.}
	\label{fig:samples}
\end{figure*}

We constructed the first ATP estimation dataset. The samples used in the dataset are tumour organoids, including colon cancer, lung cancer, nasopharyngeal cancer, ovarian cancer, and pancreatic cancer. All images were captured from organoids cultured in 384-well plates, and the corresponding ATPs were measured by ATP bioluminescence after imaging. The organoid microscopic images were acquired using the Cytation 5 imaging system, which images all wells in a plate using a uniform setting. The resolution of the captured image is $1992 \times 1992$, and the corresponding physical size is $3.45mm \times 3.45mm$. All organoids covered three distinct stages: culture stage, pre-dose and post-dose. The drugs and drug concentrations added to different wells are different, making the ATP vary in different images. We collected 1678 microscopic images and found that the range of ATP values measured by ATP bioluminescence was large, ranging from 0 to 400,000. In addition, the samples are distributed unevenly; see Figure \ref{fig:samples}. The ATPs corresponding to these images are mainly distributed in the interval of 0-200000,  and most samples are concentrated in 0-100000. Only very few samples are distributed from 200000 to 400000.

\subsection{Proposed Methodology}
To estimate ATP from microscopic images accurately, we have to solve two other challenges: (1) Organoid images are full of ATP-irrelevant targets. Directly using supervised learning to learn image-to-value mapping is easy to make the model learns redundant features. The slight perturbations in the image can make model predictions unstable. (2) The numerical range of ATP is extremely large, and the samples are unevenly distributed. This dilemma may make the model optimization unstable and inaccurate in predicting extreme samples when DCNNs are trained with limited data. To address challenge (1), we introduce a multi-instance learning mechanism that treats organoid images as bags and objects in the images as instances \cite{carbonneau2018multiple}. The instances determine the label of the bag, and the perturbation of redundant instances does not produce subversive effects on the prediction of the bag. To address challenge (2), we introduce a multi-task learning strategy, treating the problem of mapping organoid images to organoid ATP values as a combined optimization problem of regression and classification. Through multi-task training, the generalization of features can be effectively improved, and satisfying results can also be achieved on a small dataset.

\begin{figure*}[t]
	\centering
	\subfloat[Construct instances by rule]{\includegraphics[width=0.48\textwidth]{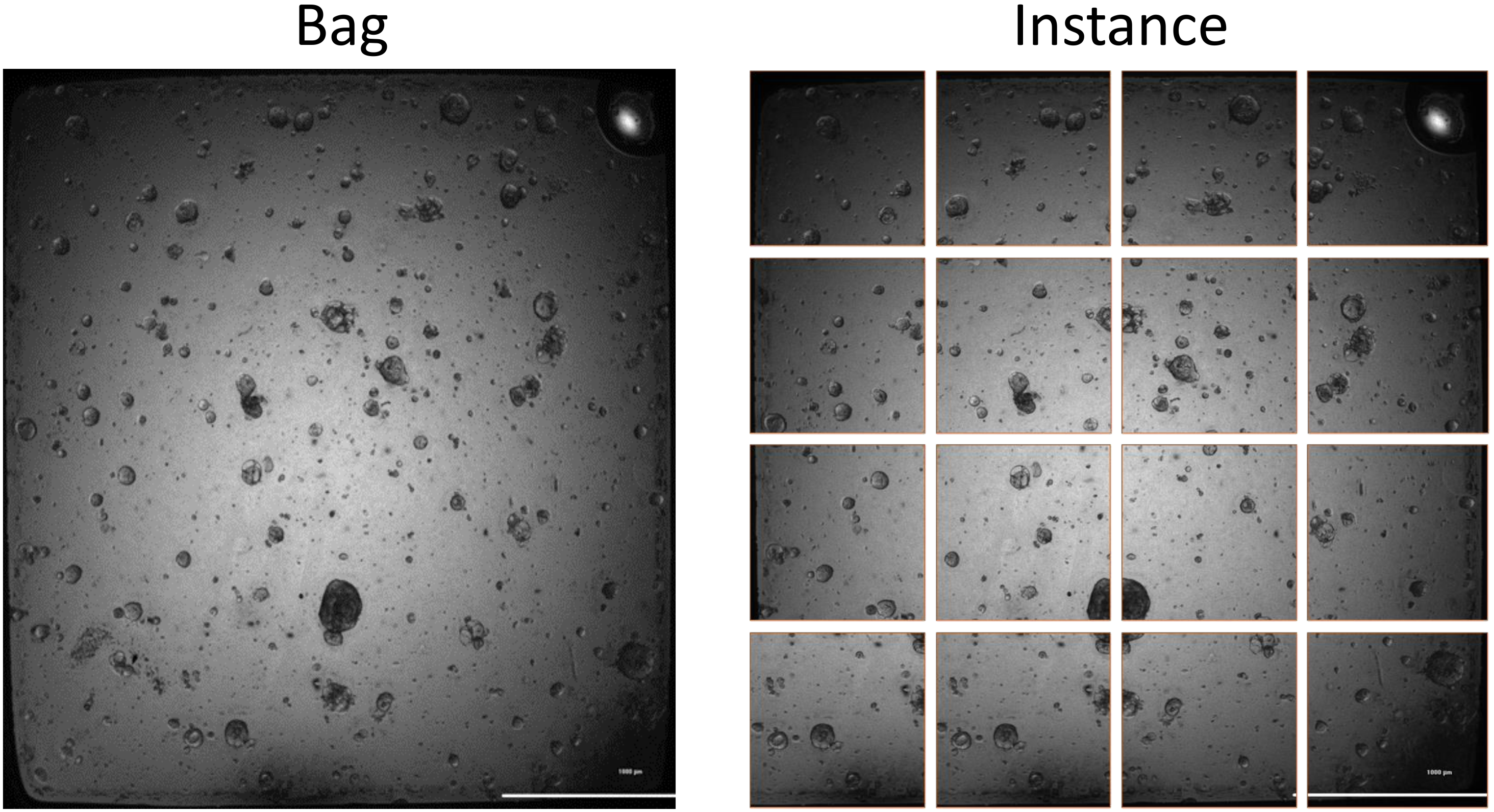}
		\label{fig:gene_instance:a}}
	\subfloat[Construct instances by model]{\includegraphics[width=0.48\textwidth]{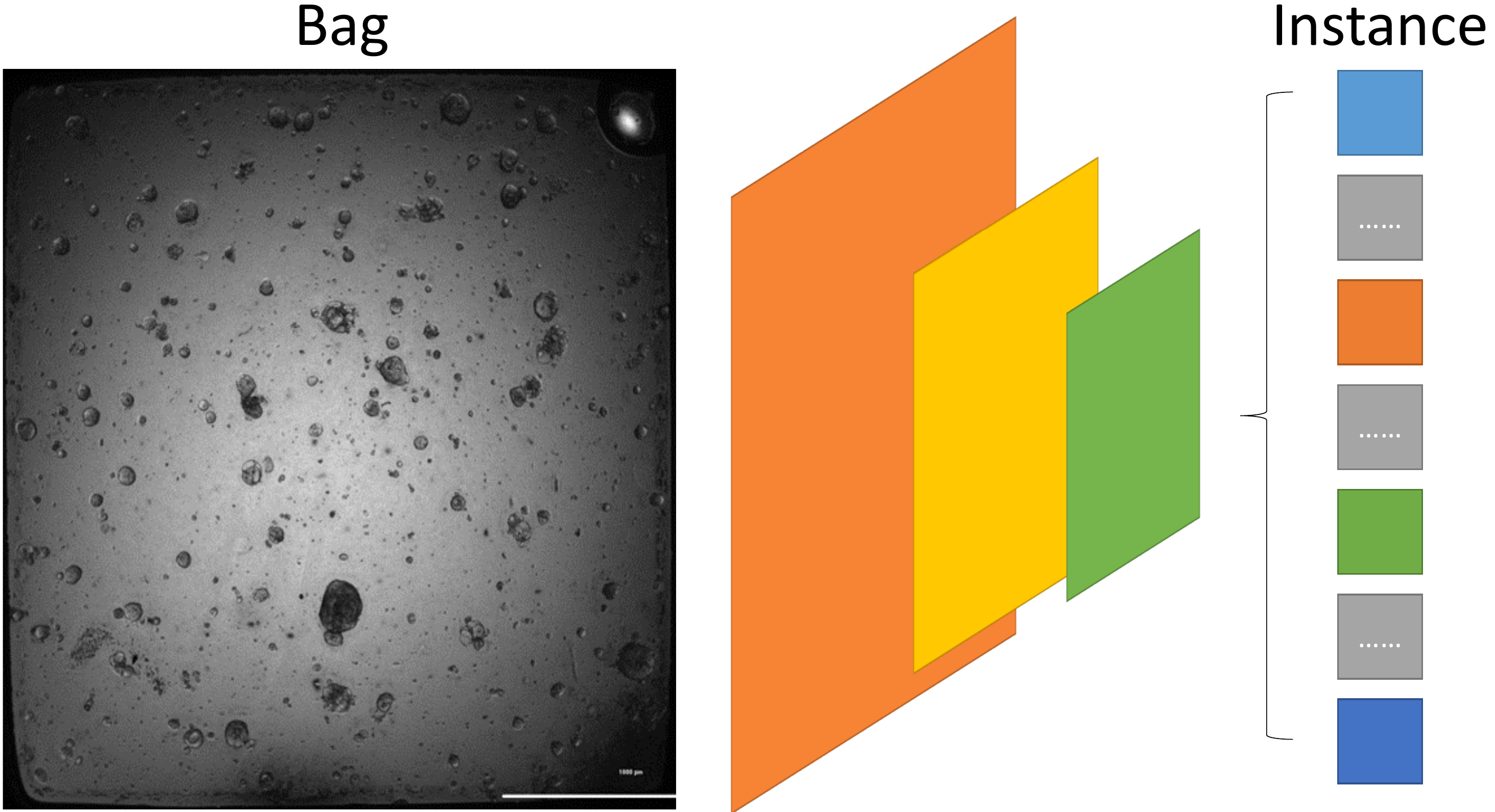}
		\label{fig:gene_instance:b}}
	\caption{Instance construction method.}
	\label{fig:gene_instance}
\end{figure*}

One of the critical factors affecting the performance of multi-instance learning is the construction strategy of instances. We propose two instances constructions methods: rule-based instance and learning-based instance. The rule-based instance is constructed according to rules, which usually requires defining instances in advance, such as cropping the image into image patches and performing organoid detection. Each image patch is regarded as an instance, as shown in Figure \ref{fig:gene_instance:a}. The learning-based instance constructs instances without manually defining instance generation rules and constructs instances by learning, as shown in Figure \ref{fig:gene_instance:b}. We propose two different architectures: MeshIns-ATP and DeepIns-ATP, based on these two instance construction schemes. The two models share other settings except for the instance construction schemes.

\begin{figure*}[t]
	\centering
	\includegraphics[width=\textwidth]{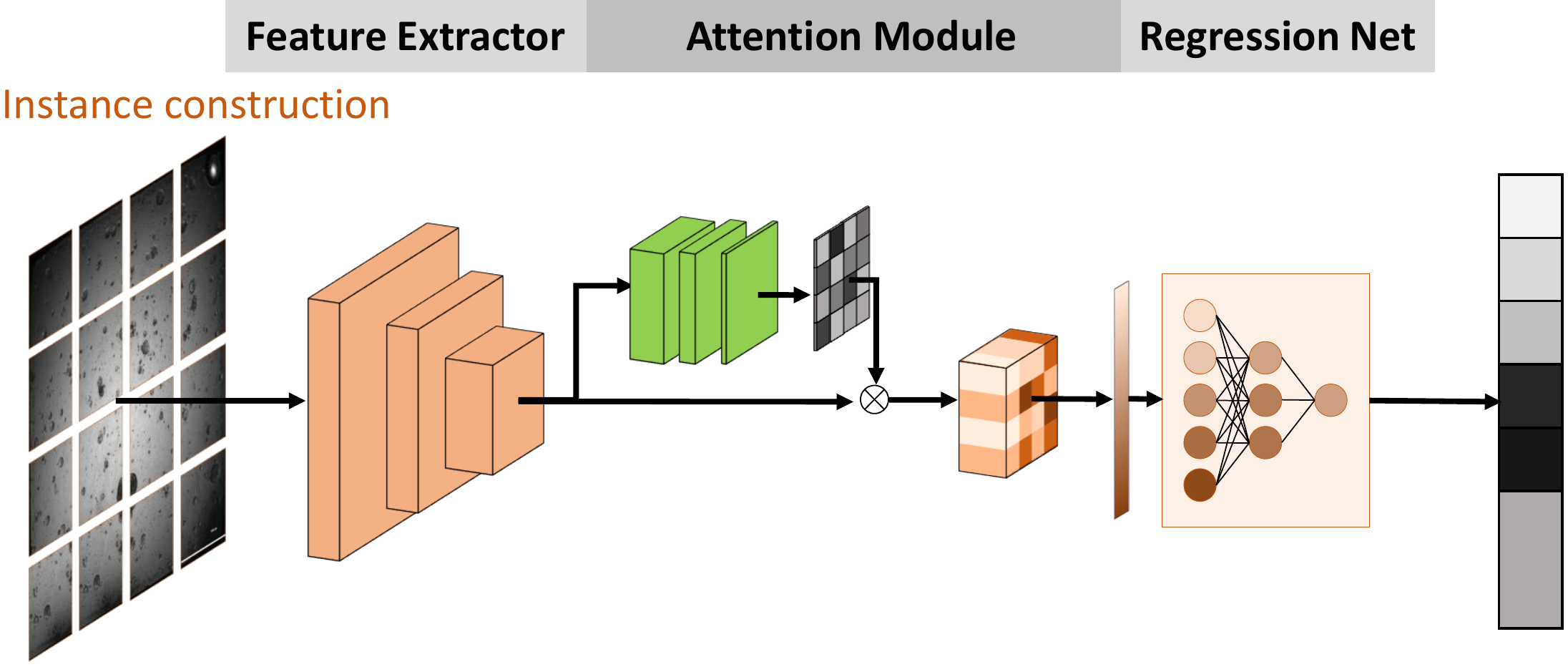}
	\caption{Architecture of MeshIns-ATP.}\label{fig:framework_1}
\end{figure*}

\begin{figure*}[t]
	\centering
	\includegraphics[width=\textwidth]{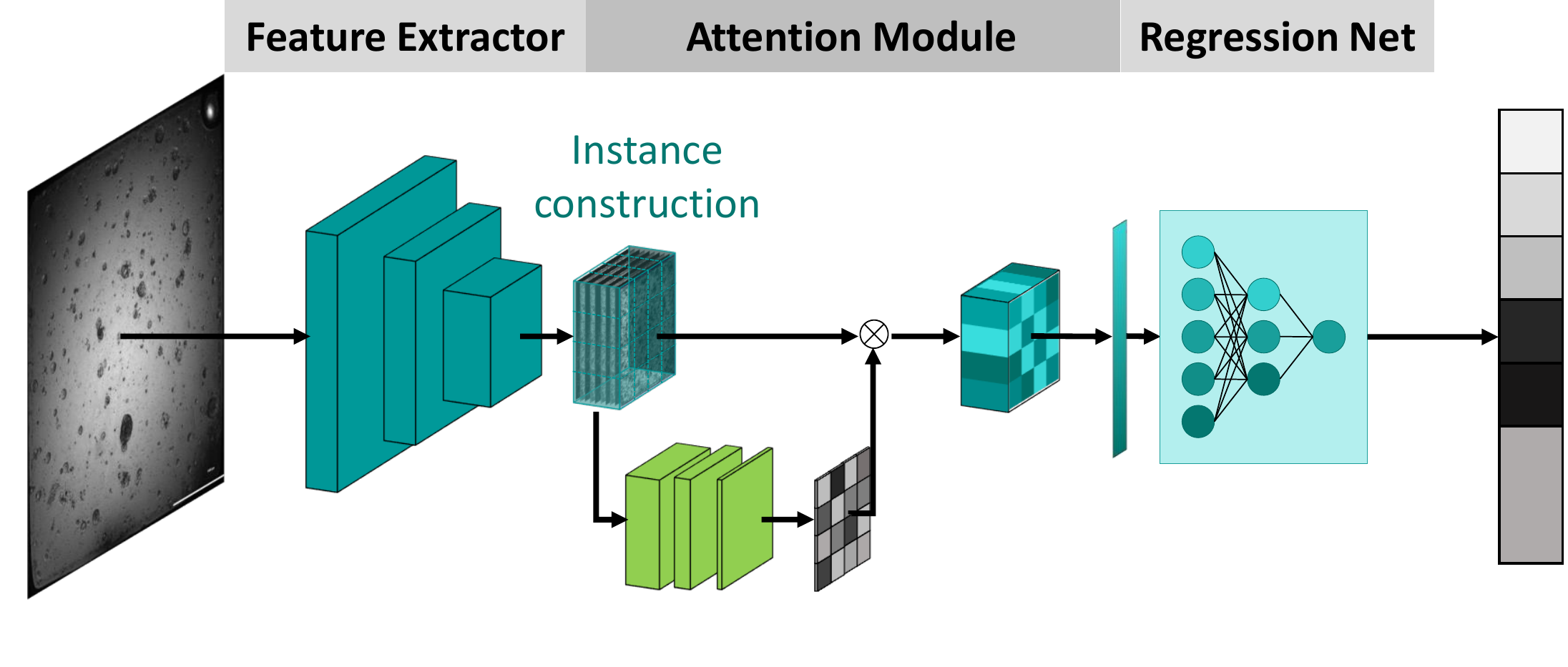}
	\caption{Architecture of DeepIns-ATP.}\label{fig:framework_2}
\end{figure*}

The Ins-ATP proposed in this paper consists of a CNN-based feature extractor, an attention module, and a regression network. The detailed network architectures are shown in Figure \ref{fig:framework_1} and Figure \ref{fig:framework_2}. The feature extractor extracts representations from the organoid image that are important for ATP estimation. The attention module weights the features of the instances in the bag according to their importance and achieves feature aggregation. The regression network maps the aggregated features to the corresponding ATP values of the images.

\subsubsection{CNN-Based Feature Extractor}

Convolutional Neural Networks (CNNs) are widely used in various computer vision tasks due to their superior performance in image feature extraction \cite{resnet,vgg,inception,googlenet}. The feature extractor of Ins-ATP used in this paper also uses CNN as an essential component. In the back-propagation during optimization, the gradient vanished quickly as the network layers increased. ResNet introduces residual connections, effectively alleviating the vanishing gradient problem \cite{resnet}. Therefore, the feature extractor in our proposed Ins-ATP also introduces the residual connection. All instances share the feature extractor. We use two feature extraction schemes here. First, construct the instances and then extract features for all instances one by one; Second, extract features from the entire image and then construct the instances. These two schemes correspond to the above two instance construction strategies, respectively.

\subsubsection{Attention Module}

Multi-instance learning takes a bag as the training unit. A bag is composed of several instances, and all instances jointly influence the prediction result of the bag. In the organoid ATP estimation task, many ATP-unrelated objects exist in the organoid image, such as impurities, vacuoles, etc. All these objects can be regarded as instances. Due to differences between instances, each instance's contribution to the bag varies. The attention mechanism is an popular method in natural language processing and then gradually extended to the field of computer vision \cite{fu2019dual,vaswani2017attention}. The attention mechanism mimics the human visual mechanism to focus on key locations in the image. 

Therefore, we introduce an attention mechanism to focus on instances crucial for ATP estimation. In general, the key instances to the prediction get higher weights. Here, we define the attention mechanism as Eq. (\ref{attention}). In the Eq. (\ref{attention}), $z$ represents the aggregated feature, $h$ represents the instance feature, $V$ represents the trainable parameters, and $a_n$ represents the instance weight. Through the attention mechanism, we get the final prediction depending on the contribution of each instance.

\begin{equation}
	\begin{aligned}
		z=&\sum_{n=1}^Na_{n}h_{n} \\
		\text{ where }\;a_n=&\frac{\exp \left\{ w^{\top}\tanh \left( V h_{n}^{\top} \right) \right\}}{\sum_{j=1}^b{\exp}\left\{ w^{\top}\tanh \left( V h_{j}^{\top} \right) \right\}}
	\end{aligned}
	\label{attention}
\end{equation}

\subsubsection{Regression Network}

The regression network is a fully connected network that maps the aggregated features to the corresponding ATP values of the images. Since the annotations range from 0 to 400000, direct optimization with ATP may lead to unstable training. So we encode the ATP with the following steps. First, we separate the range of 0 - 400000 into several bins by the bin width $R_{bin}$, then divide the ATP by the bin width $R_{bin}$. Second, binary-coded the integer part of the output to obtain the integer representation of ATP, and take the fractional part as the decimal representation of ATP. Taking the sample with ATP of 239420 as an example, the bin width is set to 20000, and the intermediate result is $239420/20000=11.971$. Take the integer part $11$ and convert it to binary representation $1011$; the fractional part is $0.971$. The encoded value is represented by a 5-dimensional vector $\{1, 0, 1, 1, 0.971\}$.

\subsubsection{Loss Function}

The mathematical notations are as follows: $\mathbb{F}(X,\theta)$ represents the organoid ATP estimation model, Ins-ATP, proposed in this paper, and $\theta$ is the parameters of Ins-ATP. We feed $N_b$ training samples as a batch to the model, $\{(X_1,Y_1), (X_2,Y_2), ..., (X_b,Y_b)\}$. $X_i$ represents the input organoid image, $Y_i$represents the ATP value corresponding to the input image, and $B$ represents the vector dimension after ATP encoding. After the above encoding, $Y_i$ can be encoded $\hat{Y}_{i,j}$, where i represents the index number of the sample, and $j$ represents the $j-th$ component of the encoded representation. To stabilize the optimization process and improve the training efficiency, we propose an auxiliary loss for training in a multi-task learning manner.

\begin{figure*}
	\centering
	\subfloat[Weight changed by decay factor $w$]{\includegraphics[width=0.48\textwidth]{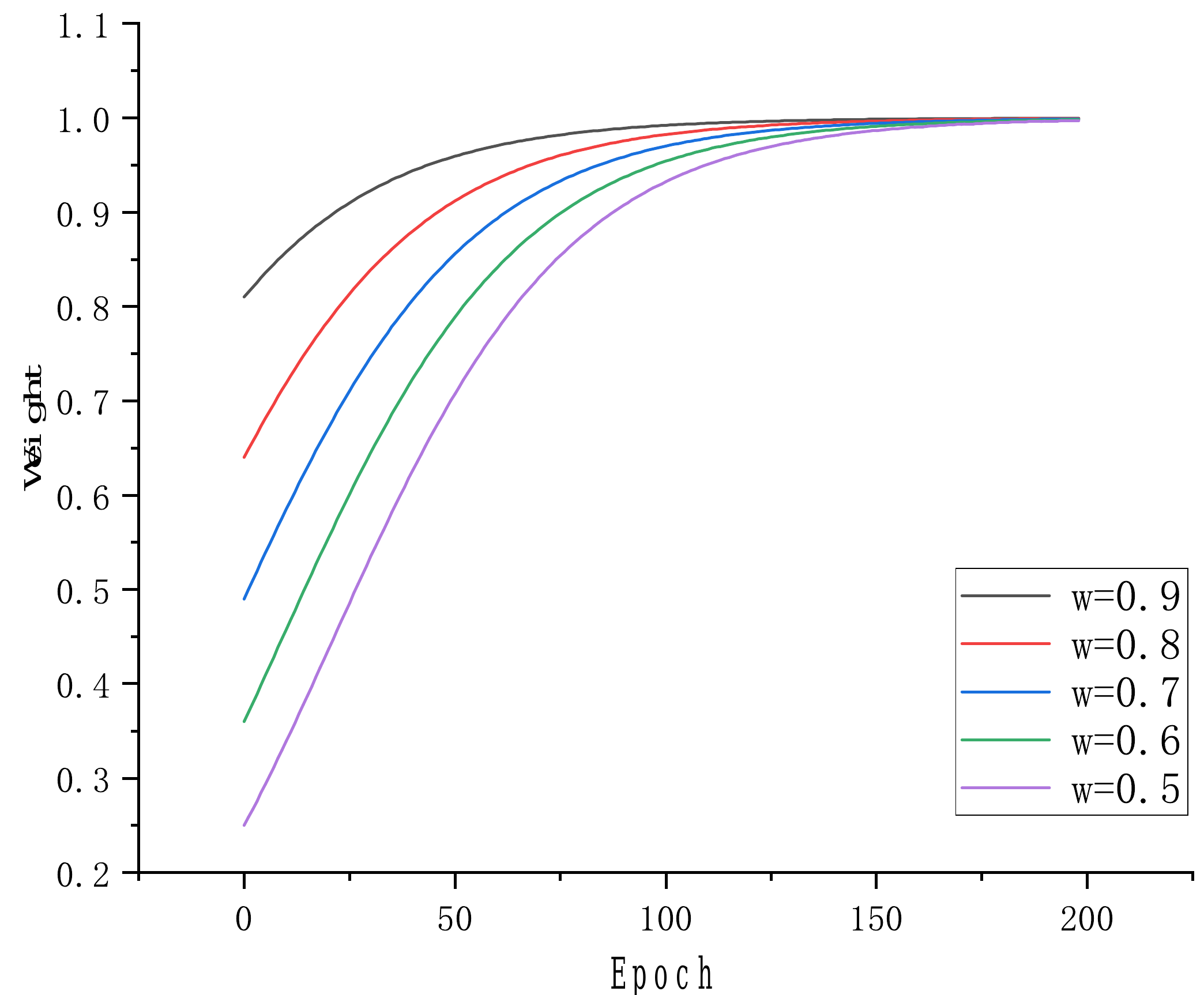}
		\label{fig:5:a}}
	\subfloat[Weight changed by encoding position $i$]{\includegraphics[width=0.48\textwidth]{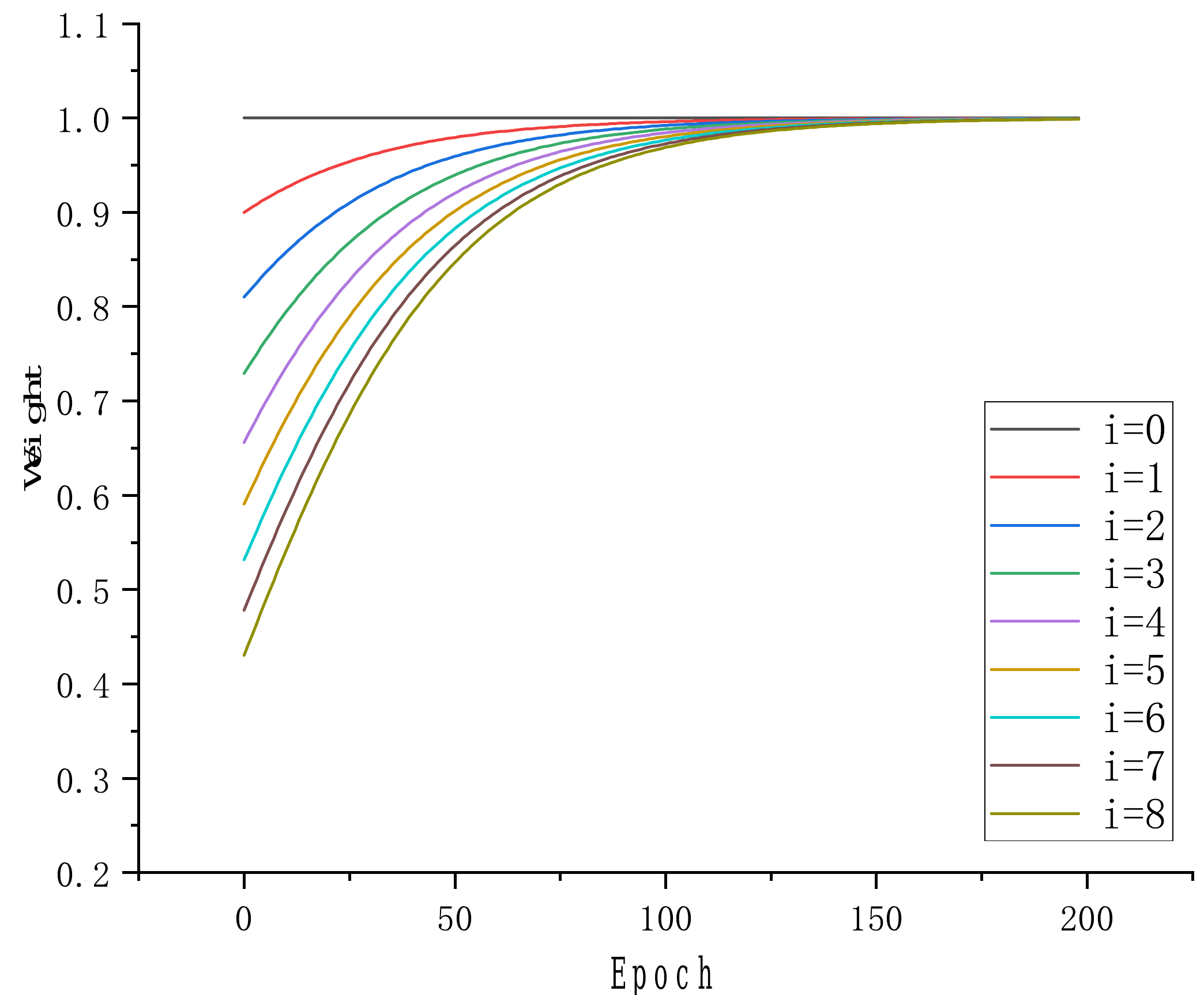}
		\label{fig:5:b}}
	\caption{Weights for different encoding position.}
	\label{fig:add_weight}
\end{figure*}

The loss function consists of two terms (Eq. (\ref{fomu:A_loss})): the first term is the ATP encoding loss (Eq. (\ref{fomu:part1})), which consists of a binary cross-entropy loss for the integer part of the ATP encoding and a mean squared error loss for the fractional part; the second term is the mean squared error loss for normalized ATP (Eq. (\ref{fomu:part2})). Considering that the different elements in the ATP-encoded vector have different contributions to the predicted value, a weighting mechanism is introduced here, see Eq. (\ref{fomu:weight}). where $i$ represents the ground truth coding dimension index, $w$ represents the decay factor, and $epoch$ represents the current training epoch. The weight of the prediction vector decreases bit by bit from left to right, and the model focuses on optimizing the high position of ATP encoding in the early stage. With the increase of the training epoch, the weight of each position is gradually balanced. It can be seen from Figure \ref{fig:add_weight} that with the increase of the number of the training epoch, the weight of each element in the ATP encoding gradually approaches 1. Figure \ref{fig:5:a} shows that by adjusting the decay factor $w$, the weight difference of each element at the beginning of training can be adjusted. The smaller the $w$, the greater the weight difference. Figure \ref{fig:5:b} shows that after the decay factor $w$ is determined, the weight difference caused by the loss of each element of the vector, the higher the component position, the higher the initial weight. 
In the Eq. (\ref{fomu:part2}), we normalize the ATP, map it to the range of $0 - 1$, and then use the mean square error to calculate the mean square loss for the normalized predicted value and the ground truth.

\begin{equation}
	\mathcal{L}=\alpha\mathcal{L}_{cls}+(1-\alpha)\mathcal{L}_{reg}
	\label{fomu:A_loss}
\end{equation}

\begin{equation}
	\begin{aligned}
		\mathcal{L}_{cls}=&\frac{1}{NB}\sum_{i=1}^N( \sum_{j=1}^{B-1}(\mathcal{L}_{bce}(\mathbb{F}(X_i,\theta)_j,\hat{Y}_{i,j})\times W_j)\\ +&\sum_{j=B-1}^{B}(\mathcal{L}_{mse}(\mathbb{F}(X_i,\theta)_j,\hat{Y}_{i,j})\times W_j)))
	\end{aligned}
	\label{fomu:part1}
\end{equation}

\begin{equation}
	\mathcal{L}_{reg}=\frac{1}{N}\sum_{i=1}^N \mathcal{L}_{mse}(\mathbb{F}(X_i,\theta),Y_i)
	\label{fomu:part2}
\end{equation}

\begin{equation}
	\begin{array}{c}
		W[i]= (\frac{1}{1+e^{(ln(\frac{1}{w}-1)-epoch/30)}})^i
	\end{array}
	\label{fomu:weight}
\end{equation}

\begin{equation}
	\begin{array}{c}
		\mathcal{L}_{bce}(x, y)=y_{i} \log x+\left(1-y\right) \log \left(1-x\right)
	\end{array}
	%	\label{fomu:loss1}
\end{equation}

\begin{equation}
	\begin{array}{c}
		\mathcal{L}_{mse}(x, y)=\left(x-y\right)^{2}
	\end{array}
	%	\label{fomu:loss2}
\end{equation}

\subsection{Optimization Strategy}

Due to the unevenness in the number of samples distributed in different bins of ATP, direct optimization by random sampling may lead to poor performance in samples distributed in specific bins of ATP. Therefore, we design a sampling strategy to ensure that the number of training data distributed in each bin remains balanced. Taking the maximum value of ATP as $ATP_{max}$, and regarding it as an example, the detailed process is as follows: (1) Set the bin width $R_{bin}$, and the number to separating bins is $ATP_{max}//R_{bin}$; (2) Suppose the training batch is $B$, then sample $B//(ATP_{max}//R_{bin})$ images uniformly from each bin; (3) Combine the samples in each bin to form the training samples for each iteration. After sampling, the number of samples participating in training is the same in each bin. The training data and the validation data are divided from the dataset randomly after shuffling.

To enhance the stability of the training, we perform image augmentation on the input image, including random horizontal and vertical flipping, overall random pixel value perturbation, and random rotation. On the one hand, the features learned by the network are robust to image flipping and brightness changes, and on the other hand, a large number of data are randomly constructed to increase the number of training samples.

The proposed Ins-ATP is implemented using the Pytorch~\cite{paszke2019pytorch} deep learning framework. The weight decay factor in the loss function is set to 0.9. The model is optimized on an Nvidia GTX 3090 GPU using the Adam optimizer, and the batch is set to 15. The learning rate is set to 0.002, which decays 0.1 every 10 epochs. We train Ins-ATP 200 epochs in total. All training data needs to be preprocessed, and only images covering the well are retained. The cropped image patch is resampled to a specific size and normalized before inputting to Ins-ATP.

\section{Experiments}\label{}

\subsection{Experimental Setup}

To validate the performance of Ins-ATP, we collected additional 335 test samples, all of which were determined by ATP bioluminescence. These samples are divided into 67 groups, and each group contains 5 samples. These samples include data from the same type of organoid with different drugs added, data from the same type of organoid sample with different concentration drugs added, and data from the same type of organoid with different drug combinations added. We design a series of experiments to verify the performance of the proposed algorithm. First, to compare predictions of Ins-ATP with the results of ATP bioluminescence, we used the mean absolute error as a metric (as shown in Eq. (\ref{fomu:MAE})). Second, to compare the correlation between the ATP value and the Ins-ATP’s evaluation at different concentrations of the drug, we use the Pearson correlation coefficient as a measure (see Eq. (\ref{fomu:pearson})).

\begin{equation}
	MAE=\frac{1}{n} \sum_{i=0}^{n}\left|\mathbb{F}(X_i,\theta)-Y_{i}\right|
	\label{fomu:MAE}
\end{equation}

\begin{equation}
	\rho = \frac{\sum_{i=0}^{n}\left(X_{i}-\bar{X}\right)\left(Y_{i}-\bar{Y}\right)}{\sqrt{\sum_{i=0}^{n}\left(X_{i}-\bar{X}\right)^{2} \sum_{i=0}^{n}\left(Y_{i}-\bar{Y}\right)^{2}}}
	\label{fomu:pearson}
\end{equation}

Firstly, the proposed Ins-ATP is compared with the traditional supervised learning methods to verify the advantages of multi-instance learning in the automatic organoid ATP estimation task. These traditional methods feed the whole image into the feature extraction network and map the learnt features directly to ATP values. We compared the performance of some popular network architectures in computer vision tasks, including VGG \cite{vgg}, ResNet \cite{resnet}, GoogLeNet \cite{googlenet}, and Inception V4 \cite{inception}.

Secondly, The parameter settings in image acquisition affect the image acquisition time. The higher the magnification of the image acquisition, the higher the resolution of the final image and the longer the acquisition takes. We should minimize the image acquisition time to alleviate the impact on organoid growth. The correlation between image resolution and model prediction performance is analyzed in our experiments. The input images are resampled at different scales, such as $128 \times 128$, $192 \times 192$, $256 \times 256$, $320 \times 320$, $384 \times 384$, $448 \times 448$, and $512 \times 512$. We retrained these models by these resampled images and analysed the trend of model performance with image resolution.

\begin{figure*}
	\centering
	\subfloat[Add feature]{\includegraphics[width=0.48\textwidth]{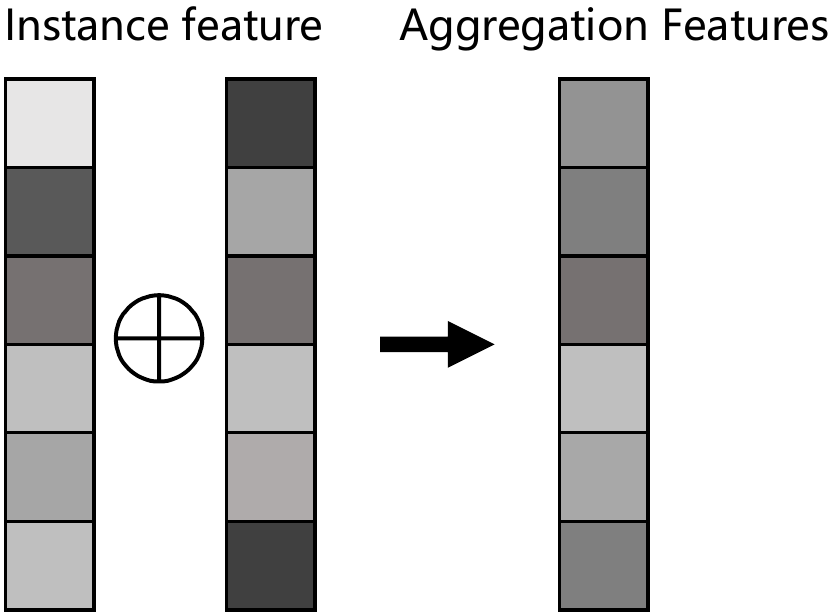}
		\label{fig:agg:a}}
	\subfloat[Concatenate feature]{\includegraphics[width=0.48\textwidth]{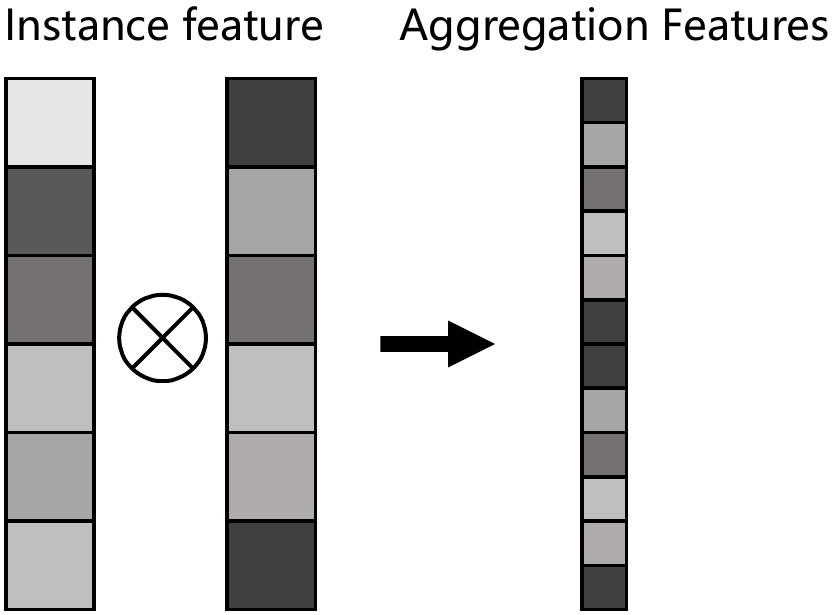}
		\label{fig:agg:b}}
	\caption{Feature aggregation strategies.}
	\label{fig:agg}
\end{figure*}

Finally, we designed ablation experiments to validate Ins-ATP proposed in this paper. Since factors such as instance generation method, instance feature aggregation strategy, and instance resampling strategy in multi-instance learning have significant effects on performance, we explore the effects of these settings on the performance of Ins-ATP. We compare the performance of the model for different instance generation methods and different numbers of instances. In addition, to investigate the impact of different feature aggregation strategies \cite{dai2021attentional}, the performance of the Ins-ATP when different feature aggregation strategies are introduced is also compared, as shown in Figure \ref{fig:agg}.

\subsection{Results}

\textbf{Comparison between multi-instance learning and traditional supervised learning}

\begin{figure*}
	\centering
	\includegraphics[width=\textwidth]{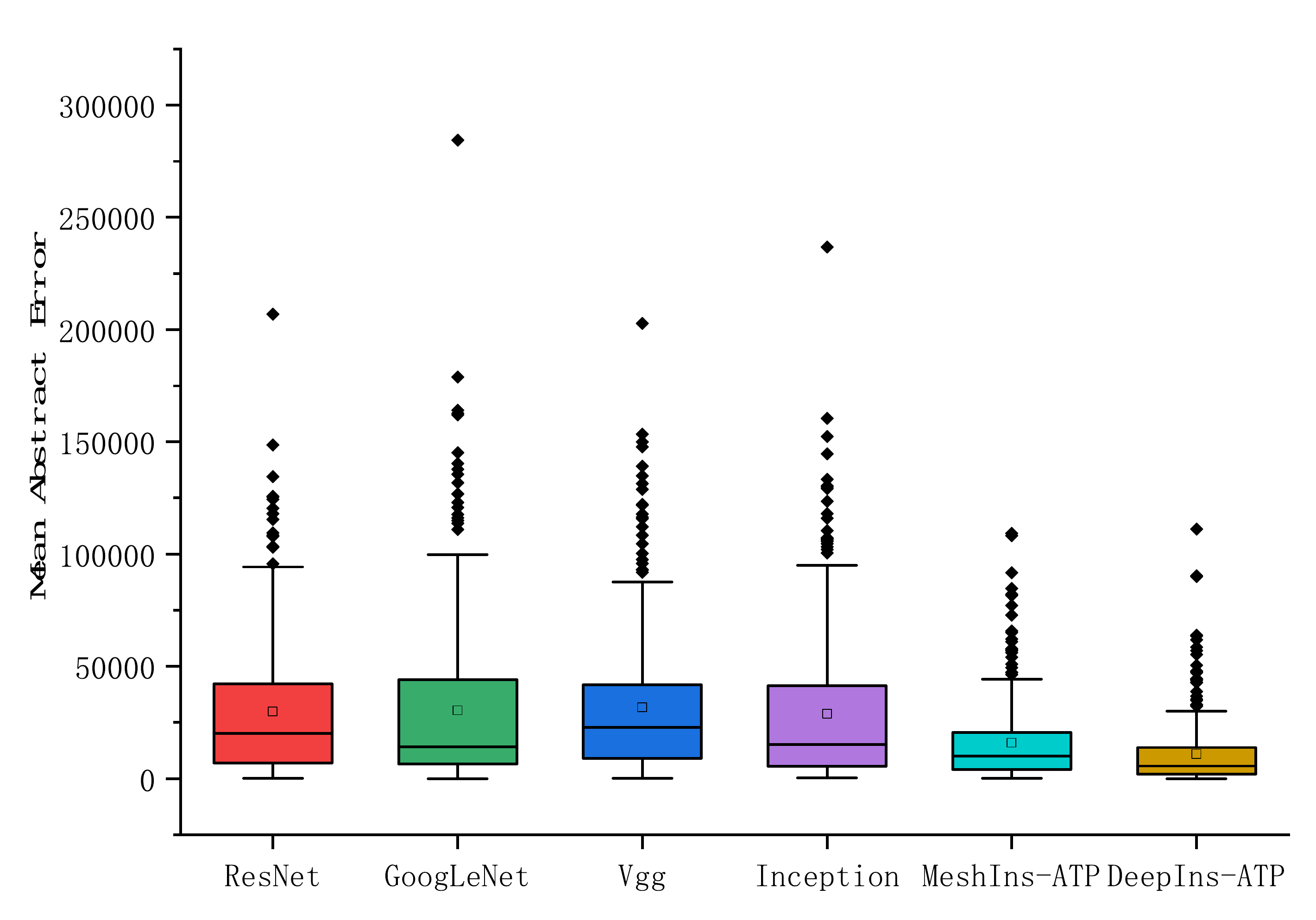}
	\caption{Prediction error of different models.}\label{fig:box}
\end{figure*}

\begin{figure*}
	\centering
	\subfloat[ResNet]{\includegraphics[width=0.3\textwidth]{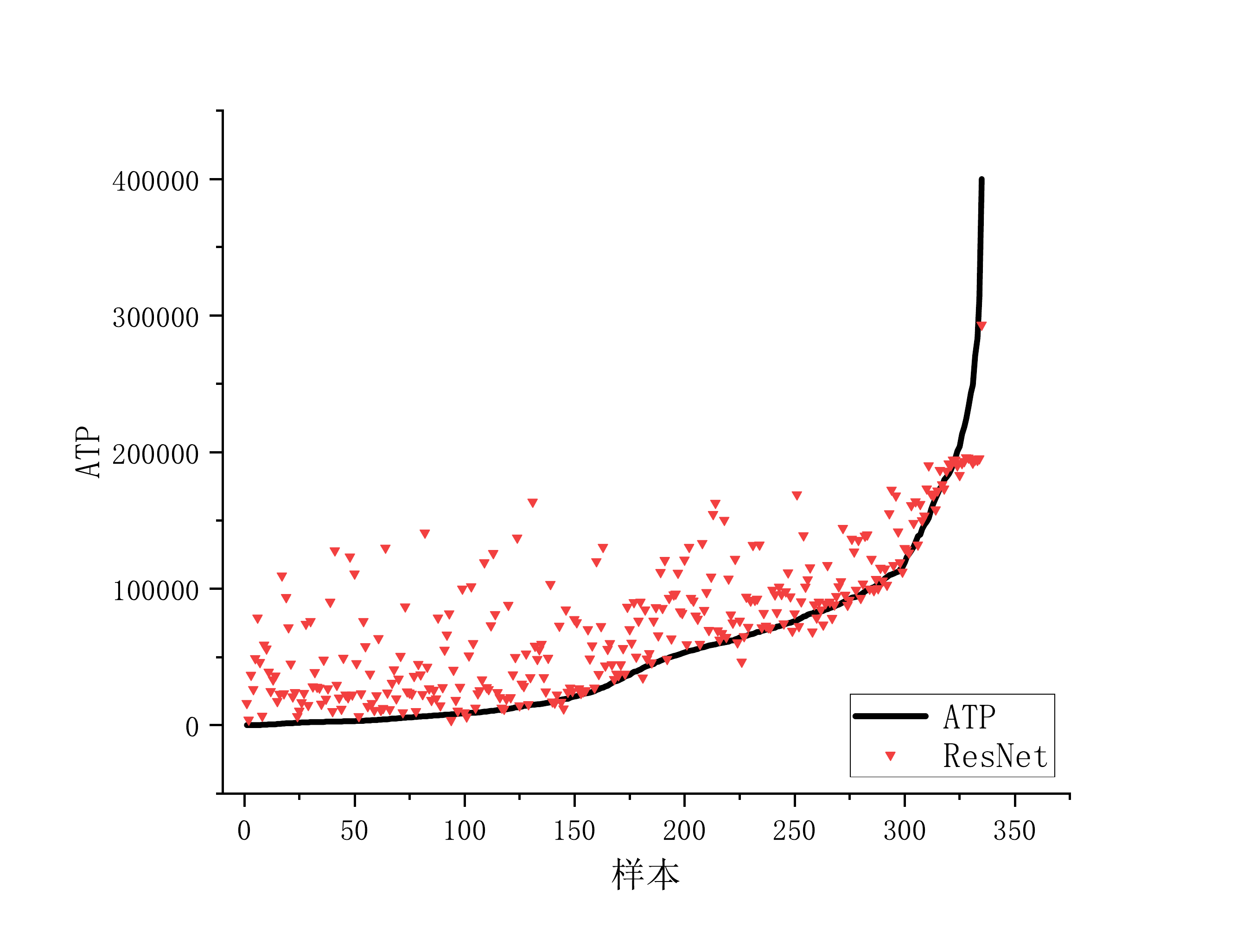}
		\label{fig:method_compare:a}}
	\subfloat[GoogLeNet]{\includegraphics[width=0.3\textwidth]{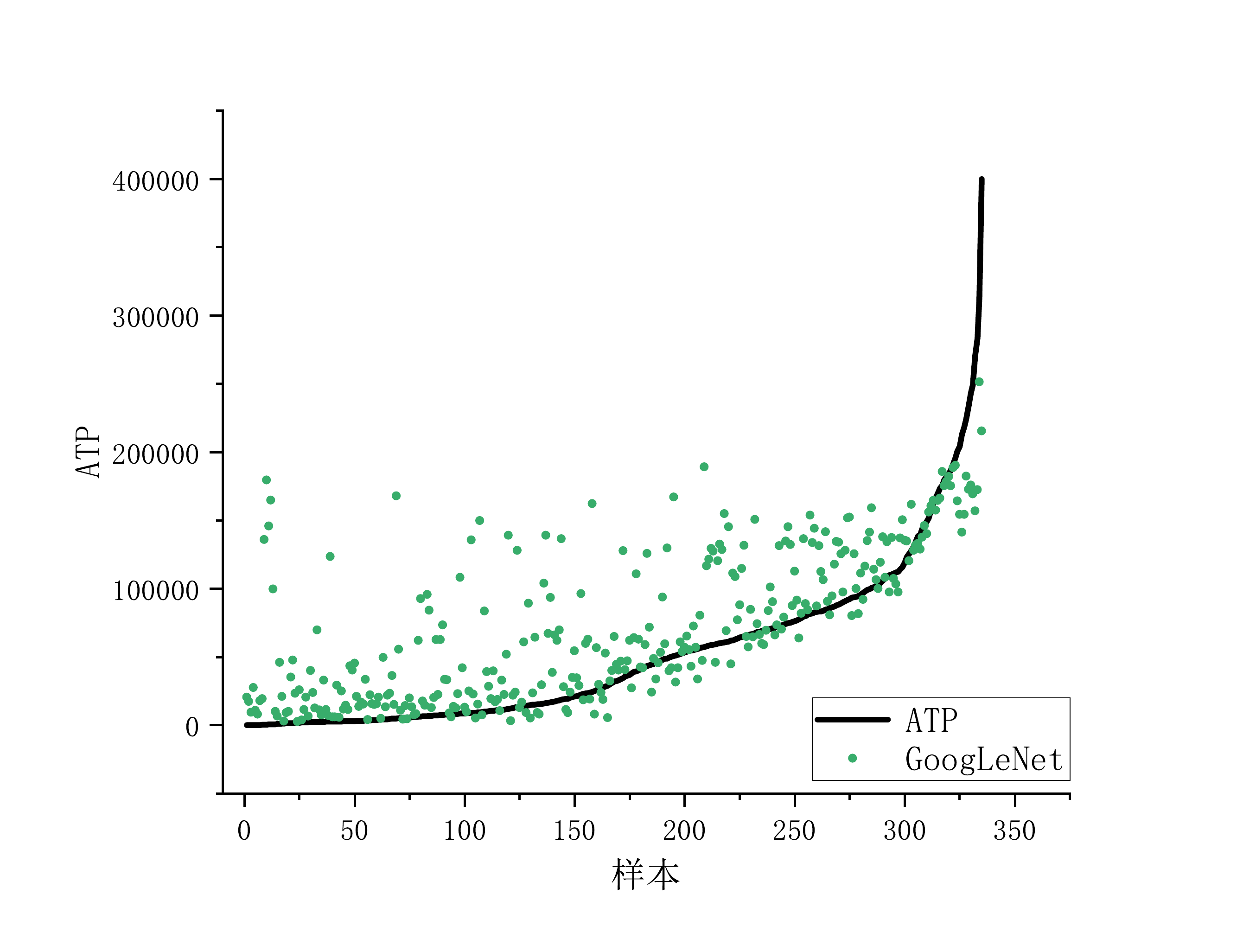}
		\label{fig:method_compare:c}}
	\subfloat[Vgg]{\includegraphics[width=0.3\textwidth]{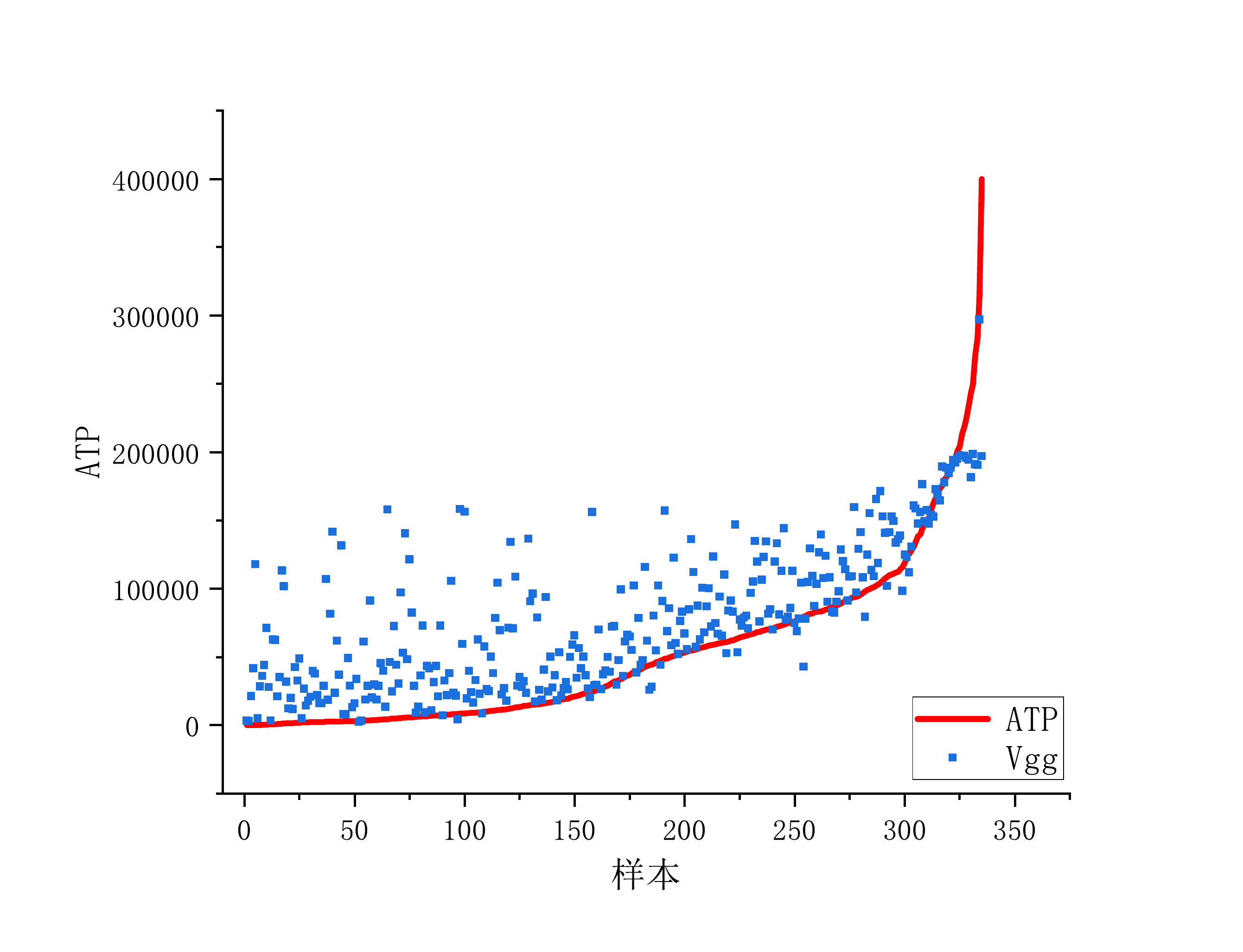}
		\label{fig:method_compare:d}}\\
	\subfloat[Inception Net]{\includegraphics[width=0.3\textwidth]{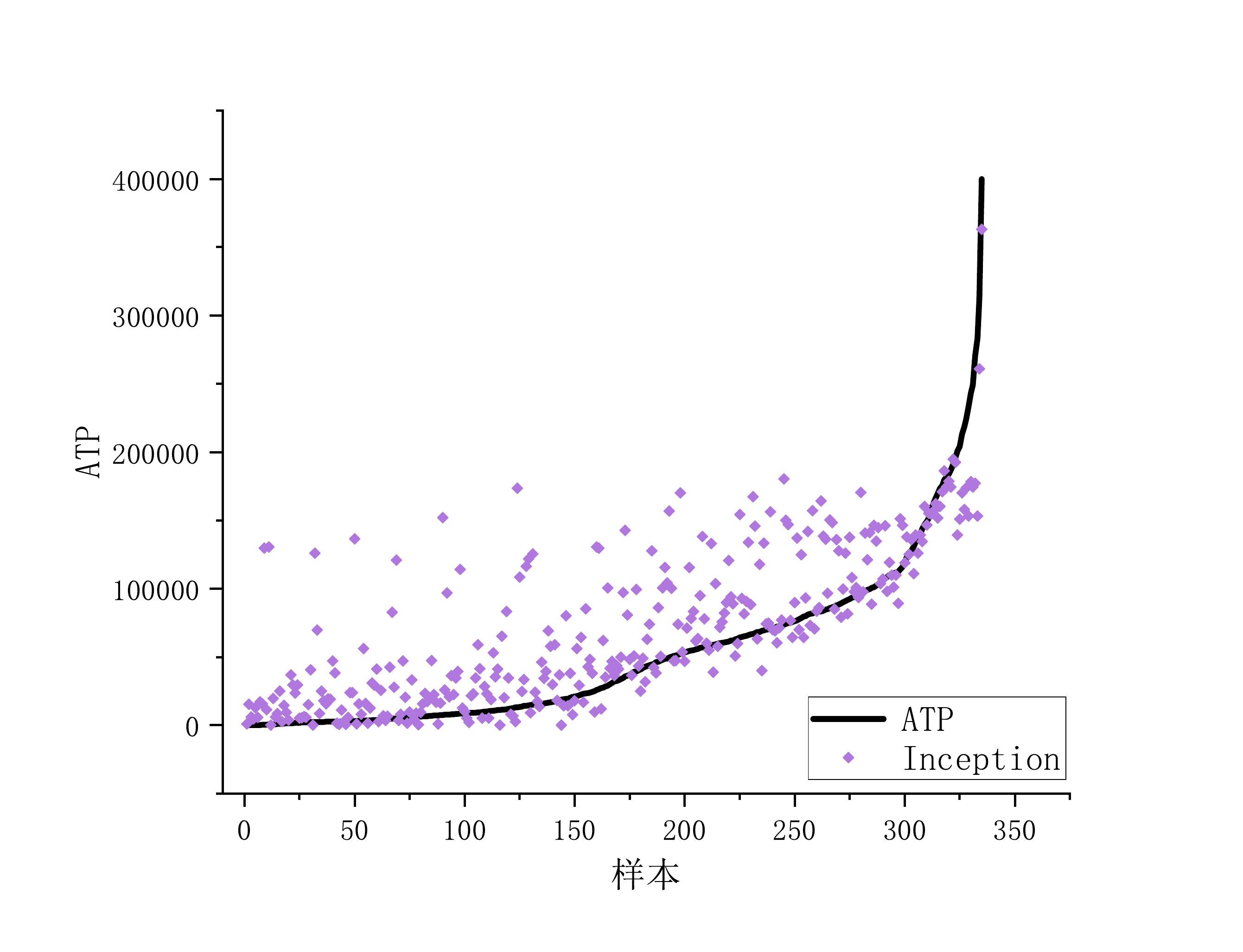}
		\label{fig:method_compare:b}}
	\subfloat[MeshIns-ATP]{\includegraphics[width=0.3\textwidth]{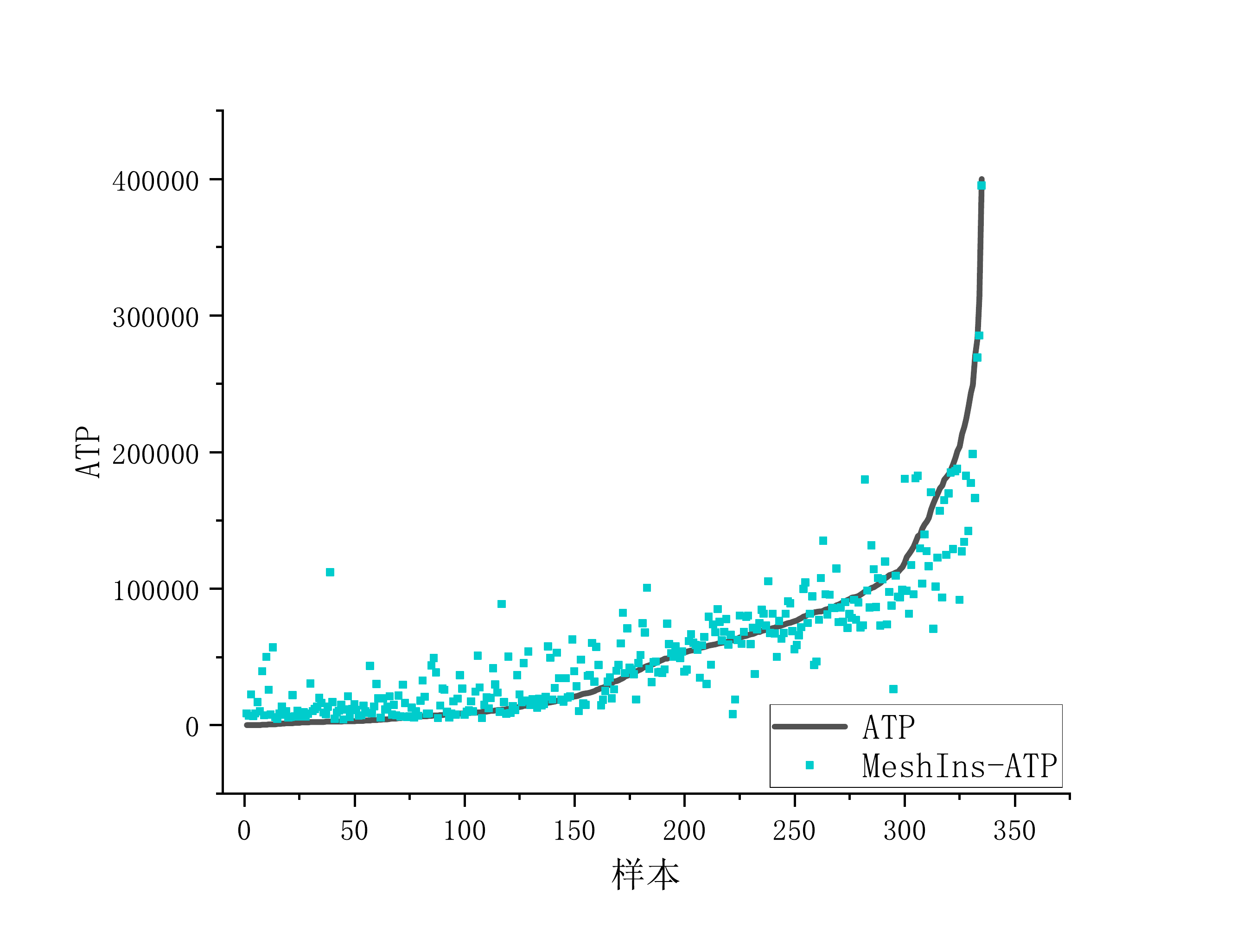}
		\label{fig:method_compare:e}}
	\subfloat[DeepIns-ATP]{\includegraphics[width=0.3\textwidth]{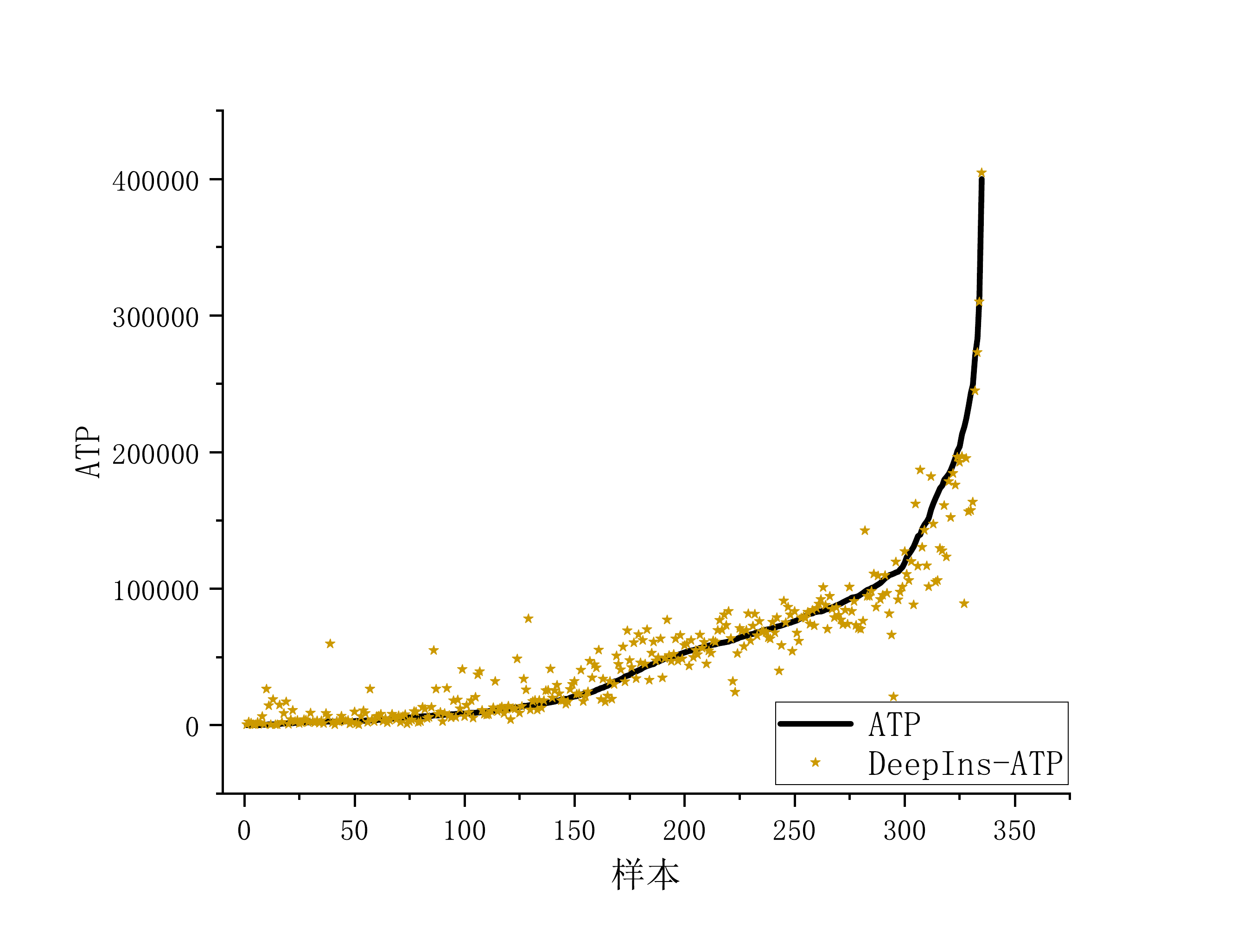}
		\label{fig:method_compare:f}}	
	\caption{Difference between the prediction of the model and the ground-truth.}
	\label{fig:method_compare}
\end{figure*}

We first compare the Ins-ATP with the mainstream supervised learning methods. The models involved in the comparison include VGG \cite{vgg}, ResNet \cite{resnet}, GoogleNet \cite{googlenet}, and Inception Net V4 \cite{inception}. The above four models all use the models provided in the Pytorch framework and are initialized with pre-trained parameters on ImageNet \cite{krizhevsky2012imagenet}. All models are trained using the same training data. The experimental results are shown in Table \ref{tab:eval}. It can be found that the two methods (MeshIns-ATP and DeepIns-ATP) based on multi-instance learning proposed in this paper outperform the above four mainstream models in both evaluation metrics. It can be seen from Figure \ref{fig:box} that the prediction errors of the above four models have a large variance, which indicates that these models have significant differences in the prediction accuracy of different samples. The variance of the prediction errors of Ins-ATP was kept low. As shown in Figure \ref{fig:method_compare}, the above four models have poor prediction performance in the interval where the real sample distribution is sparse. The multi-instance learning method, MeshIns-ATP and DeepIns-ATP, proposed in this paper can maintain a balanced performance in the entire dataset.

\begin{table*}[]
	\caption{Performance comparison of different model.}
	\label{tab:eval}
	\centering
	\begin{tabular}{|cc|c|c|}
		\hline
		\multicolumn{2}{|c|}{Method}                                    & Mean Absolute Error (MAE)         & Pearson Correlation Coefficient ($\rho$)           \\ \hline
		\multicolumn{1}{|c|}{\multirow{4}{*}{Traditional supervised learning}} & ResNet \cite{resnet}      & 25166          & 0.593          \\ \cline{2-4} 
		\multicolumn{1}{|c|}{}                       & GoogleNet \cite{googlenet}   & 28654          & 0.595          \\ \cline{2-4} 
		\multicolumn{1}{|c|}{}                       & VGG \cite{vgg}         & 30470          & 0.499          \\ \cline{2-4} 
		\multicolumn{1}{|c|}{}                       & Inception V4 \cite{inception}& 26271          & 0.611          \\ \hline
		\multicolumn{1}{|c|}{\multirow{2}{*}{Multi-instance learning}} & MeshIns-ATP  & 14784          & 0.792          \\ \cline{2-4} 
		\multicolumn{1}{|c|}{}                       & DeepIns-ATP  & \textbf{10149} & \textbf{0.898} \\ \hline
	\end{tabular}
\end{table*}

\begin{figure*}
	\centering
	\subfloat[Losses in training set]{\includegraphics[width=0.49\textwidth]{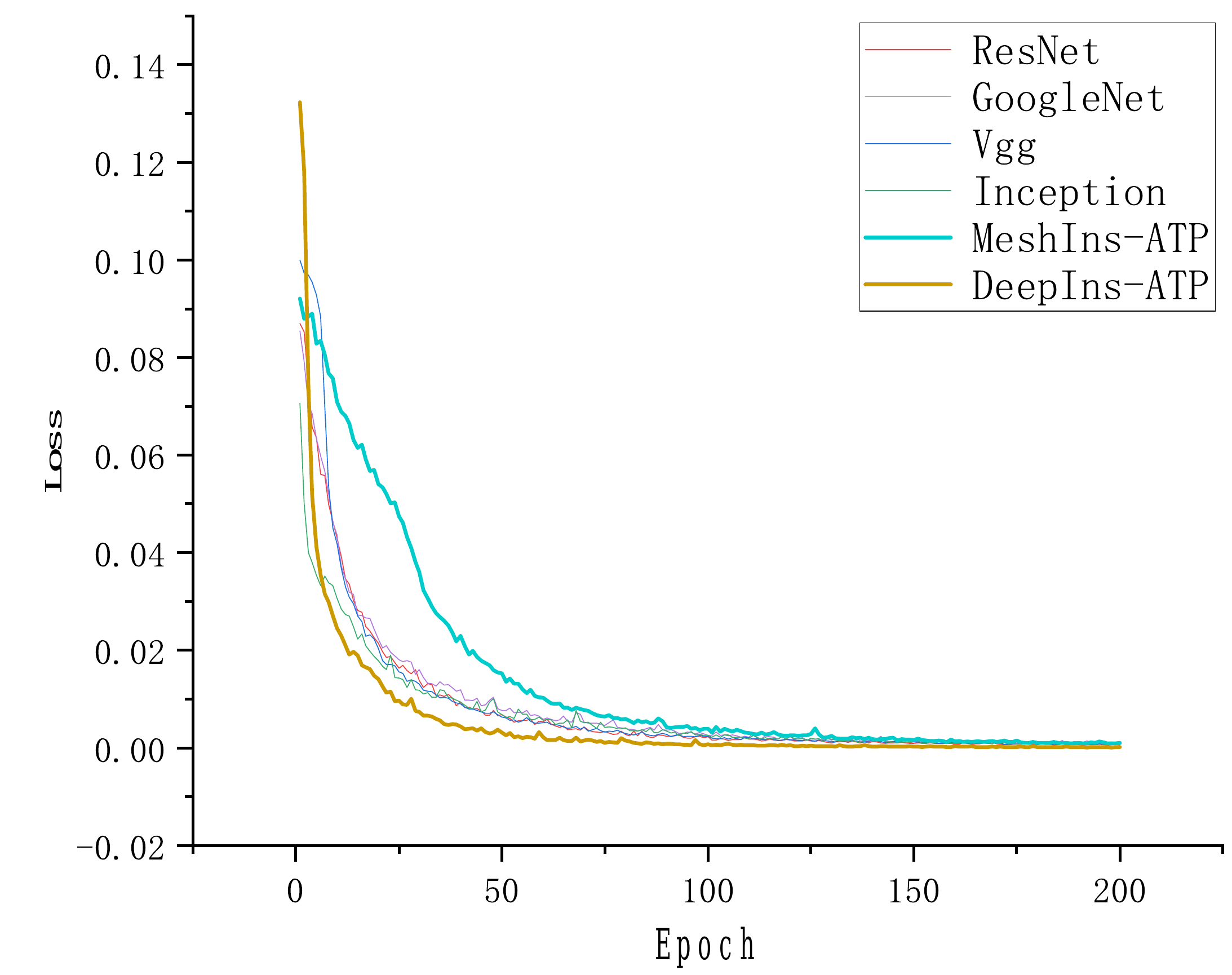}
		\label{fig:loss_eval:a}}
	\subfloat[MAE in validation set]{\includegraphics[width=0.49\textwidth]{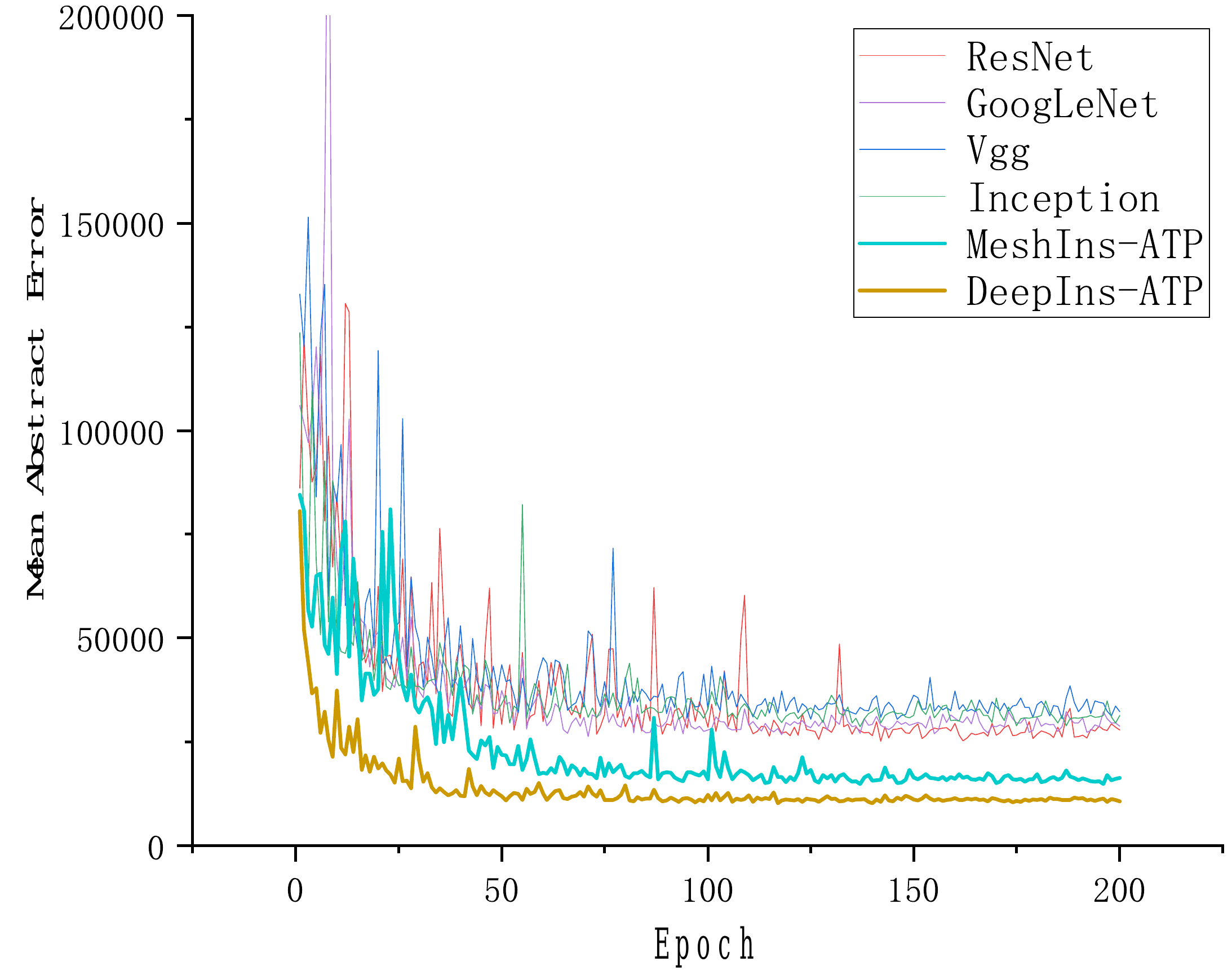}
		\label{fig:loss_eval:b}}
	\caption{Visualization of training process.}
	\label{fig:loss_eval}
\end{figure*}

We visualize the optimization of all models, as shown in Figure \ref{fig:loss_eval:a}. During the optimization of DeepIns-ATP, the overall downward trend of the loss curve remains stable, and the convergence speed is fast. In addition, as shown in Figure \ref{fig:loss_eval:b}, the test set also maintains a steady decline without violent jitter. The other four models have serious jitter on the test set, indicating that the Ins-ATP also maintain good generalization performance on the test set.

\textbf{Performance comparison at different image resolutions}

\begin{figure}
	\centering
	\includegraphics[width=0.5\textwidth]{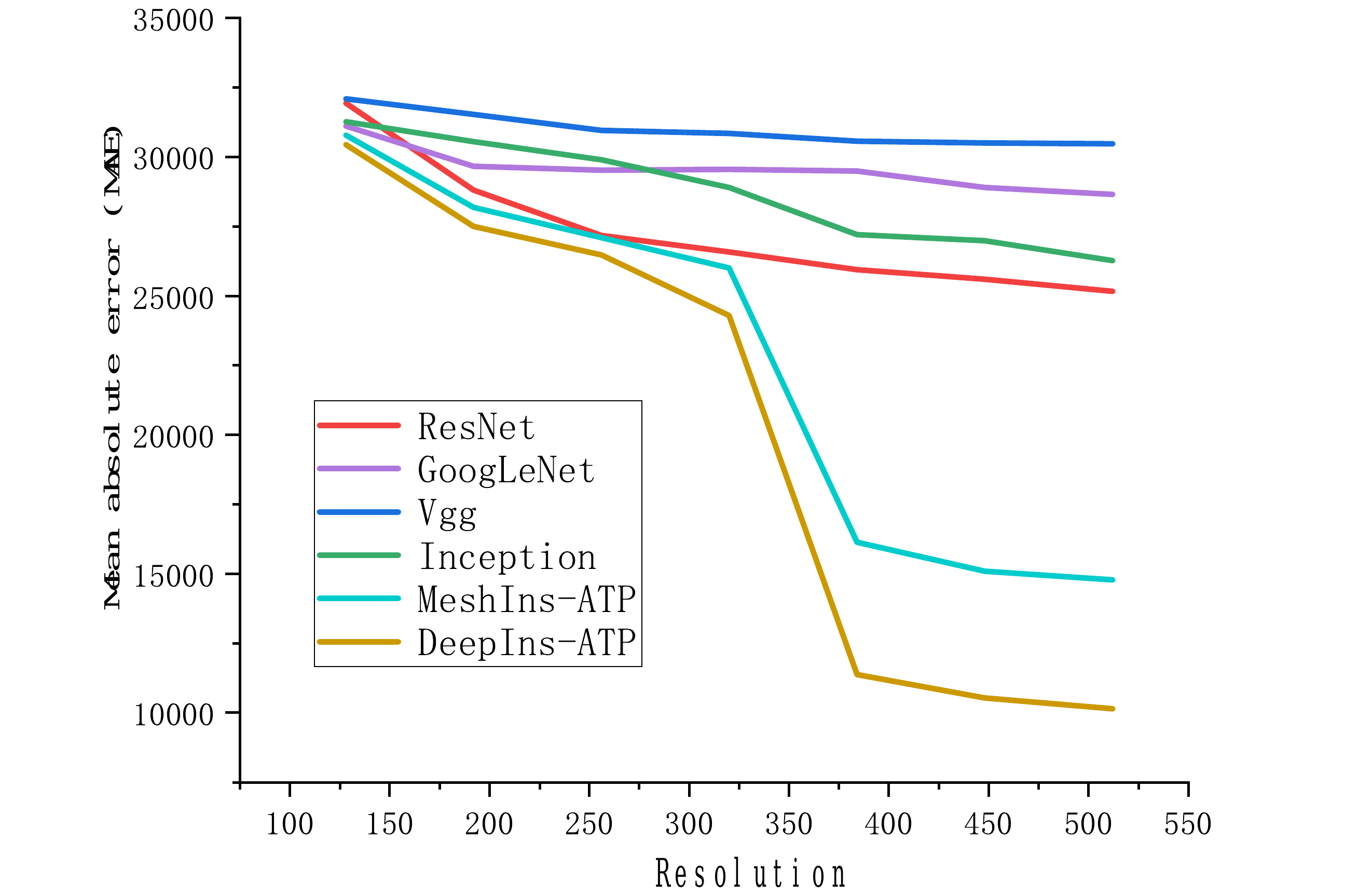}
	\caption{Trend of model prediction error with image resolution.}\label{fig:resolution}
\end{figure}

Depending on the practical needs, microscopic images are acquired using different settings, such as magnification, number of stitching, etc. To analyze the performance of each model at different resolution images and to obtain the best resolution for training the models, we train the models with microscopic images at different resolutions. The experimental results are shown in Table \ref{tab:resulution_mae}. It can be seen that the prediction error of each model decreases with increasing resolution from Figure \ref{fig:resolution}. When the resolution is low, all models have high prediction errors because of image quality problems. As the resolution increases, the performance of each model gradually increases, especially the MeahIns-ATP and DeepIns-ATP based on multi-instance learning proposed in this paper, whose performance improves significantly as the resolution increases.

\begin{table*}[]
	\caption{Performance comparison at different image resolutions.}
	\label{tab:resulution_mae}
	\centering
	\begin{tabular}{|c|c|c|c|c|c|c|}
		\hline
		Resolutions     & ResNet \cite{resnet} & GoogLeNet \cite{googlenet} & VGG \cite{vgg} & Inception \cite{inception}  & MeshIns-ATP & DeepIns-ATP \\ \hline
		128$\times$128 & 31933 & 31102 & 32083 & 31266 & 30790 & 30447 \\ \hline
		192$\times$192 & 28804 & 29656 & 31534 & 30551 & 28184 & 27504 \\ \hline
		256$\times$256 & 27178 & 29519 & 30948 & 29897 & 27096 & 26468 \\ \hline
		320$\times$320 & 26581 & 29547 & 30848 & 28898 & 26006 & 24290 \\ \hline
		384$\times$384 & 25948 & 29486 & 30560 & 27204 & 16130 & 11380 \\ \hline
		448$\times$448 & 25597 & 28900 & 30500 & 26979 & 15088 & 10542 \\ \hline
		512$\times$512 & 25166 & 28654 & 30470 & 26271 & 14784 & \textbf{10149} \\ \hline
	\end{tabular}
\end{table*}

\textbf{Ablation study}

Finally, we designed the ablation experiments of each module in Ins-ATP, including whether to introduce ATP encoding, whether to use resampling for images, different instance generation methods and different feature fusion methods. First, we divided the experiments into two groups according to the instance generation method: dividing the instances by grid before inputting the feature extractor and automatically generating the instances after inputting the feature extractor. Secondly, we conducted experiments for the three feature fusion methods. Three feature fusion methods were tried: (1) summing all instance features directly; (2) stitching all instance features and combining them into a higher dimensional feature; (3) weighting all example features to sum by the attention mechanism. Finally, an experimental comparison is made on whether to introduce ATP coding or not. 

In the model training process, two training strategies are set in this paper according to whether to resample the images or not to explore the practicality of the resampling method for data with uneven sample distribution. The results of all the above experiments are shown in Table \ref{tab:ablation}. The experiments show that the approach of generating instances with learning is more suitable for the task of ATP estimation, and its performance is better than generating instances by mesh. Introducing attention weighting for the feature fusion approach improves the prediction performance. In addition, the introduction of sample resampling and encoding of ATP can also help in performance improvement.

\begin{table*}[]
	\caption{Performance comparison of algorithms under different settings.}
	\label{tab:ablation}
	\centering
	\begin{tabular}{|c|c|cc|cc|}
		\hline
		\multirow{2}{*}{Instance generation method} & \multirow{2}{*}{Feature fusion method} & \multicolumn{2}{c|}{Sample resampling}   & \multicolumn{2}{c|}{ATP Encoding}   \\ \cline{3-6} 
		&                         & \multicolumn{1}{c|}{True}  & False  & \multicolumn{1}{c|}{True}  & False  \\ \hline
		\multirow{3}{*}{Instance by mesh} & Summing feature                     & \multicolumn{1}{c|}{17855} & 18461 & \multicolumn{1}{c|}{17985} & 18994 \\ \cline{2-6} 
		& concatenating feature                    & \multicolumn{1}{c|}{18695} & 19152 & \multicolumn{1}{c|}{17544} & 17854 \\ \cline{2-6} 
		& weighting feature by attention                   & \multicolumn{1}{c|}{16842} & 17844 & \multicolumn{1}{c|}{15784} & 16736 \\ \hline
		\multirow{3}{*}{Instance by learning} & Summing feature                     & \multicolumn{1}{c|}{13997} & 15892 & \multicolumn{1}{c|}{12566} & 13264 \\ \cline{2-6} 
		& concatenating feature                  & \multicolumn{1}{c|}{17578} & 18542 & \multicolumn{1}{c|}{13648} & 15468 \\ \cline{2-6} 
		& weighting feature by attention                   & \multicolumn{1}{c|}{\textbf{10926}} & 13223 & \multicolumn{1}{c|}{11149} & 12429 \\ \hline
	\end{tabular}
\end{table*}

\section{Discussion}\label{}

We propose a deep learning algorithm based on multi-instance learning in this paper to solve the problem of automatic estimation of ATP from microscopic images for organoids. In addition, we propose a loss function based on ATP encoding to optimize the model. In addition, a data resampling strategy is proposed to cope with the problem of uneven distribution of samples in the ATP prediction space. The model performance is verified through a large number of experiments, while the effects of various designs in the network on the model performance are compared through ablation experiments. The model proposed in this paper is analyzed below.

The experimental comparison shows that the estimation results of the Ins-ATP outperform the traditional supervised learning method on the test set for the automatic organoid ATP estimation task. The model optimization process based on multi-instance learning is more stable compared to traditional supervised learning methods, and this stability can maintain the advantage over the test data. It is because multi-instance learning treats the whole microscopic image as a bag, and changes in individual instances within the bag do not have a disruptive effect on the prediction of the whole bag by the attention mechanism. Due to various targets in the image that are redundant to ATP estimation, supervised learning strategies have difficulty learning robust features in a small amount of data.

By comparing the input images with different resolutions, we found that different models prefer training data with different resolutions but generally show the same trend. It can be found that the higher the resolution of the training images, the better the performance of the trained models obtained. As the image resolution decreases, a large number of details in the images are lost, thus leading to a poorer discriminative ability of the model on the data. Through the experiments, it can be found that the performance improvement of the model is weak after the resolution exceeds $384 \times 384$. Therefore, through experiments, the most suitable resolution can be found for this task to achieve a compromise between efficiency and accuracy.

Through extensive comparison experiments, each module and setting proposed in this paper positively affects performance improvement. By increasing the number of examples, the performance is significantly improved. The more examples, the less dependent the model is on specific examples, the more robust the model is to perturbations of individual examples, and the more accurate predictions it can make. In addition, the performance is sensitive to the interval width setting. Since the interval width setting affects the collateral loss in multi-task learning, a small interval width increases the number of intervals divided and the discrimination between samples increases, which affects the optimization of the ATP coding loss and the model training. Therefore, several experiments are needed to try the optimal setting.

Although the method in this chapter performs better on the automatic organoid ATP estimation task, the method has some limitations. First, the method is difficult to generalize on the same task for different organoid images, and the model needs to be retrained before it is used on other types of organoid data. Second, the method in this paper is designed only for data captured from 384-well plates. The number of organoids in different sizes of well plates can vary significantly due to the different volumes of the inner wells of the plates, and the differences in the values of ATP determined by bioluminescence may be extremely large. Therefore, this method is only compatible with tumour organoid images cultured in 384-well plates.

\section{Conclusions}\label{}
In this paper, we constructed a dedicated dataset for automatic ATP estimation of tumour organoids. We proposed a deep learning network, Ins-ATP, based on multi-instance learning to achieve accurate ATP estimation for tumour organoids. Our Ins-ATP effectively achieves accurate estimation of ATP from organoids' microscopic images. The method treats organoid microscopic images as a bag consisting of multi-instances and the organoid ATP estimation problem as a bag prediction problem. 
To optimize the model and achieve regression with a wide range of values, we proposed a loss function based on ATP encoding, which maps the predicted value to a vector. We performed joint optimization by combining the cross-entropy loss and the mean squared error loss. To overcome the problem that the training is unstable due to the imbalance of training data in the ATP, we proposed a training data sampling strategy to improve the stability of the training. 
Experiments show that the method based on multi-instance learning proposed in this paper is superior to the traditional supervised learning methods in organoid ATP estimation tasks. The prediction of our Ins-ATP is in good agreement with the measurement results of ATP bioluminescence. 

\section*{Declaration of competing interest}\label{}
We would like to confirm that all authors were fully involved in the study and preparation of the manuscript. None of this work has been previously published, or been pending publication in another journal, or been under review in any other journal. None of the authors has a conflict of interest.

\section*{Acknowledgments}\label{}


\bibliographystyle{cas-model2-names}

\bibliography{ref}

%% Here goes the biography details.

\end{document}